\documentclass[journal,twoside,web]{ieeecolor}
\usepackage[table]{xcolor}
\usepackage{tmi}
\usepackage{cite}
\usepackage{amsmath,amssymb,amsfonts}
\usepackage{algorithmic}
\usepackage{graphicx}
\usepackage{textcomp}
\usepackage{contour}
\usepackage{bm}
\usepackage[normalem]{ulem}
\usepackage{float}
\usepackage{amsfonts}
\usepackage{url}
\usepackage{stmaryrd}
\usepackage{booktabs}
\usepackage{rotating}
\usepackage{multicol}
\usepackage{multirow}
\usepackage{booktabs}   
\usepackage{bbding} 

\newcommand{\ourcell}{\cellcolor{gray!10}}
\newcommand{\ci}[1]{\scriptsize{\textcolor{gray}{~($\pm #1$)}}}
\usepackage{makecell}
\usepackage{subcaption} 
\usepackage[colorlinks,linkcolor=blue,anchorcolor=blue,citecolor=blue,]
{hyperref}
\def\BibTeX{{\rm B\kern-.05em{\sc i\kern-.025em b}\kern-.08em
    T\kern-.1667em\lower.7ex\hbox{E}\kern-.125emX}}
\newcommand{\figref}[1]{Fig.~\ref{#1}}
\newcommand{\tabref}[1]{Tab.~\ref{#1}}

\newcommand{\secref}[1]{Sec.~\ref{#1}}

\def\ourmodel{\emph{Self-P2IR}}
\def\ourdata{\emph{P2I-LReg}}

\def\ie{\emph{i.e.}}
\def\eg{\emph{e.g.}}

\def\etal{{\em et al.}}

\begin{document}
\title{
Landmark-Free Preoperative-to-Intraoperative Registration in Laparoscopic Liver Resection
}

\author{Jun Zhou, Bingchen Gao, Kai Wang, Jialun Pei, \IEEEmembership{Member, IEEE}, Pheng-Ann Heng, \IEEEmembership{Senior Member, IEEE}, \\ and Jing Qin, \IEEEmembership{Senior Member, IEEE}
\thanks{Jun Zhou, Bingchen Gao, and Jing Qin are with the Center of Smart Health, School of Nursing, The Hong Kong Polytechnic University, HKSAR, China. (e-mail: zachary-jun.zhou@connect.polyu.hk;  bingchen.gao@connect.polyu.hk; harry.qin@polyu.edu.hk)}
\thanks{Jialun Pei and Pheng-Ann Heng are with the Department of Computer Science and Engineering, Pheng-Ann Heng is also with the Institute of Medical Intelligence and XR, The Chinese University of Hong Kong, HKSAR, China. (e-mail: jialunpei@cuhk.edu.hk; pheng@cse.cuhk.edu.hk)}
\thanks{Kai Wang is with the Division of Hepatobiliopancreatic Surgery, Department of General Surgery, Nanfang Hospital, Guangzhou, China. (e-mail: kaiwang@smu.edu.cn)}
\thanks{Jun Zhou and Bingchen Gao contributed equally.}
\thanks{Corresponding author: Jialun Pei.}
}

\maketitle

\begin{abstract}
Liver registration by overlaying preoperative 3D models onto intraoperative 2D frames can assist surgeons in perceiving the spatial anatomy of the liver clearly for a higher surgical success rate.
Existing registration methods rely heavily on anatomical landmark-based workflows, which encounter two major limitations: 1) ambiguous landmark definitions fail to provide efficient markers for registration; 2) insufficient integration of intraoperative liver visual information in shape deformation modeling. To address these challenges, in this paper, we propose a landmark-free preoperative-to-intraoperative registration framework utilizing effective self-supervised learning, termed \ourmodel. 
This framework transforms the conventional 3D-2D workflow into a 3D-3D registration pipeline, which is then decoupled into rigid and non-rigid registration subtasks.
\ourmodel~first introduces a feature-disentangled transformer to learn robust correspondences for recovering rigid transformations. Further, a structure-regularized deformation network is designed to adjust the preoperative model to align with the intraoperative liver surface. This network captures structural correlations through geometry similarity modeling in a low-rank transformer network. To facilitate the validation of the registration performance, we also construct an in-vivo registration dataset containing liver resection videos of 21 patients, called \emph{P2I-LReg}, which contains 346 keyframes that provide a global view of the liver together with liver mask annotations and calibrated camera intrinsic parameters. Extensive experiments and user studies on both synthetic and in-vivo datasets demonstrate the superiority and potential clinical applicability of our method. 
The code and dataset are available at \href{https://github.com/junzastar/Self-P2IR.git}{Self-P2IR}.
\end{abstract}

\begin{IEEEkeywords}
3D/2D registration, Laparoscopic Liver Resection, Self-Supervised Learning, Augmented Reality
\end{IEEEkeywords}

\section{Introduction}
\label{sec:introduction}

\begin{figure}[t!]
	\centering
	\begin{subfigure}{0.49\textwidth}
		\centering
		\includegraphics[width=\linewidth]{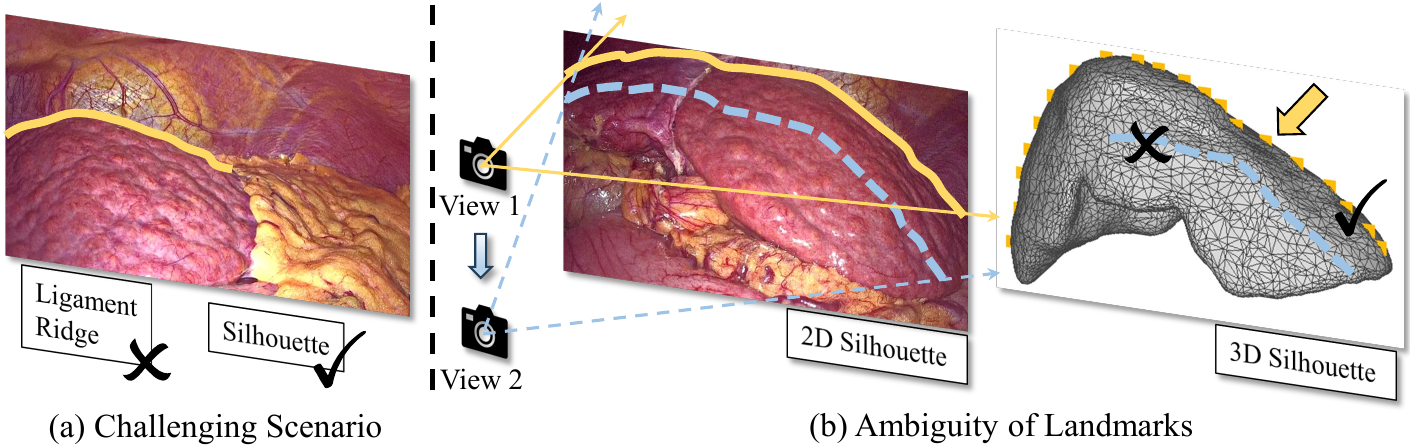}
	\end{subfigure}
	\vspace{-10pt}
	
	\begin{subfigure}{0.49\textwidth}
		\centering
		\includegraphics[width=\linewidth]{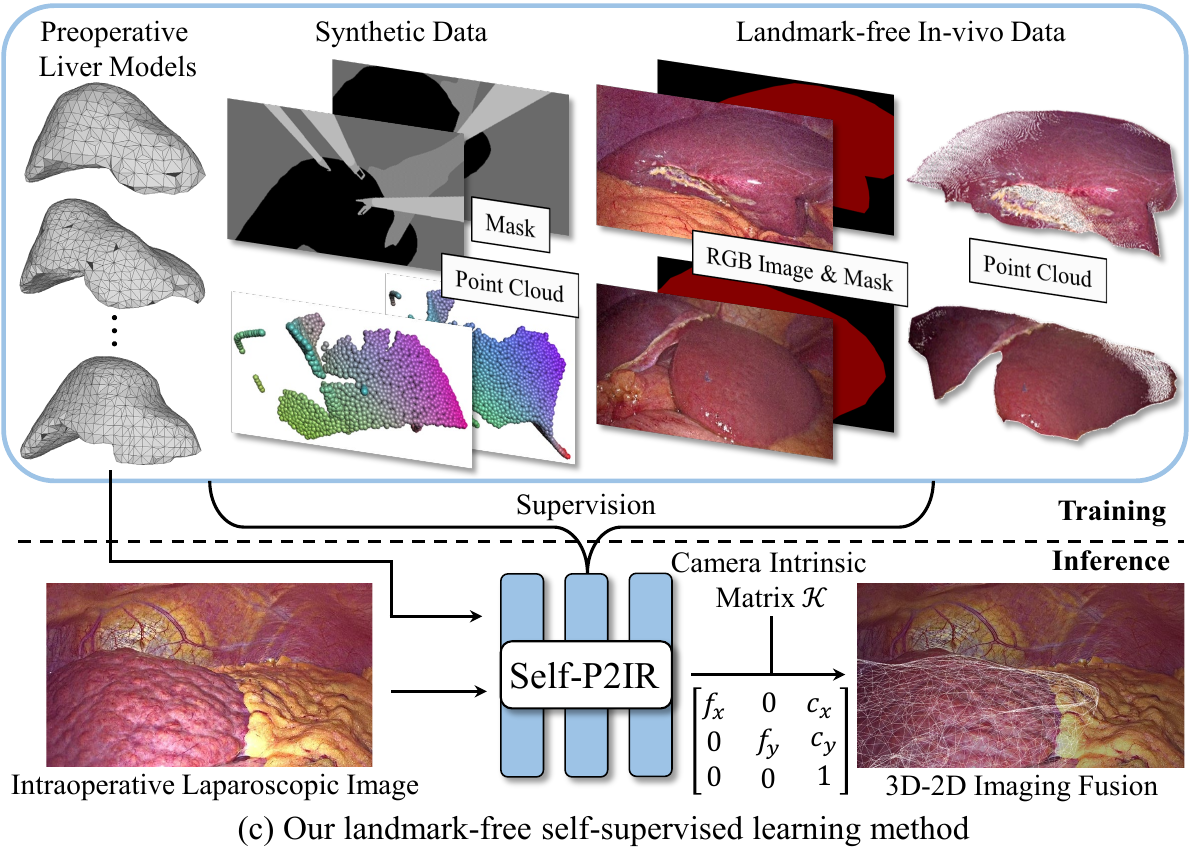}
	\end{subfigure}
	\caption{
		Illustration of (a) challenges of landmarks in laparoscopic scenarios; (b) ambiguity of landmark annotations, with 2D landmarks varying by camera view; (c) our registration method without landmarks or interactive inputs for inference.
	}
	\label{fig:teaser_2}
\end{figure}

\IEEEPARstart{A}{ugmented} reality (AR) system provides a realistic visualization of stereoscopic imaging for laparoscopic liver resection (LLR), enabling surgeons to accurately identify the internal anatomy of the liver.
Precise imaging guidance to determine the relative position between the tumor and major vessels can enhance the success of liver resection and reduce the risk of postoperative complications~\cite{ramalhinho2023value,guo2025surgical}.
The essence of the intraoperative AR-assist system is the overlay of the 3D liver model from preoperative CT or MRI with intraoperative images, \ie, 3D-2D registration.
However, the registration process in laparoscopy faces several challenges: (i) partial visibility of the liver surface caused by the limited laparoscopic view and occlusions (\eg, blood and instruments); (ii) the lack of distinctive texture features on the liver surface; and (iii) significant deformation of the liver caused by pneumoperitoneum and tool manipulation.
These challenges make preoperative and intraoperative registration still a nontrivial task~\cite{koo2022automatic,mhiri2024neural}.

Early registration methods typically involve two main steps: global rigid registration first and local iterative optimization post-processing. However, global registration requires interactive inputs depending on the surgeon’s expertise, involving manual adjustments~\cite{pelanis2021evaluation,ozgur2018preoperative}  and landmark-based semi-automatic alignment~\cite{feuerstein2008intraoperative}.
Later, some approaches~\cite{koo2017deformable,espinel2022using,koo2022automatic, pei2024land} primarily employ landmark-based matching strategy, by detecting landmarks (\eg, ridges, ligaments, and silhouettes) in both 2D intraoperative images and 3D preoperative models, and then recovering the pose parameters by solving the 3D-2D Perspective-n-Point (PnP) problem. 
Nevertheless, this landmark-based registration scheme suffers from the following limitations: 
1) \emph{Difficulty in precise landmark annotations:} Accurate correspondence between landmarks in intraoperative images and preoperative models is essential for 3D-2D registration. In clinical practice, however, liver landmarks are partially visible due to limited laparoscopic view and occlusions, particularly for ridge and ligament landmarks (see~\figref{fig:teaser_2}(a)). 
2) \emph{Ambiguity in landmark definition:} Although the silhouette of the liver is a crucial reference to assess liver pose, its definition remains ambiguous. 
The intraoperative liver silhouette from the laparoscopic view changes with the camera angle (see~\figref{fig:teaser_2}(b)). 
This variability complicates landmark matching and makes registration more difficult.
Moreover, the manual costs of landmark annotation also pose a challenge in obtaining large numbers of labels.
As a result, we intend to introduce a landmark-free preoperative-to-intraoperative registration strategy in laparoscopic liver resection.

During the deformation phase, current techniques~\cite{labrunie2023automatic, mhiri2024neural} primarily adopt training strategies based on simulated data, \ie, generating a set of deformed liver models for each patient. These methods aim to learn the mapping relationship between landmarks and deformation parameters to predict the deformation field for preoperative liver models. 
However, existing techniques rely solely on local landmark semantic features to predict global liver deformation parameters without fully leveraging the visual information of the intraoperative liver visible surface, resulting in inaccurate deformation and registration predictions in complex scenes. Additionally, due to the uncertainty of intraoperative liver deformation, existing approaches based on synthetic data exhibit considerable domain gaps that exacerbate the uncertainty in deformation field estimation.
In this regard, exploring intraoperative visual cues, \ie, 2D shape contour and 3D geometry, to model the preoperative-intraoperative geometrical correlation is essential for accurately estimating liver deformation.

To overcome these obstacles, we propose a landmark-free self-supervised framework, termed~\ourmodel, for preoperative-to-intraoperative liver registration (see~\figref{fig:teaser_2}(c)). 
Our framework converts the traditional 3D-2D registration workflow into a 3D-3D pipeline, \ie, partial-to-global point cloud registration between intraoperative reconstructed 3D models and preoperative 3D models.
\ourmodel~disentangles the registration process into two procedures: rigid registration (global alignment) and non-rigid registration (deformation). 
For rigid registration, we represent the input using a superpoint scheme to ensure computational efficiency and embrace a Feature Disentangled Transformer (FDT) to empower the semantic correlation between preoperative and intraoperative liver models.
In this way, we can leverage the corresponding confidence matrix to estimate the global registration pose parameters.
To facilitate preoperative model deformation, we propose an effective Structure-Regularized Shape Adaptation (SRSA) strategy that measures the structural similarity between preoperative and intraoperative models in order to exploit their inherent geometric characteristics. 
Moreover, \ourmodel~adaptively injects semantic features from intraoperative models into preoperative models to effectively mitigate the semantic gap.
Notably, our framework enables self-supervised training by incorporating self-contained masks and geometric cues from intraoperative liver models.
To bridge the limitations of existing registration datasets, including insufficient cases and the lack of annotations for liver 2D mask and 3D point cloud, we construct a new in-vivo dataset, termed \ourdata~for registration performance evaluation.
We conduct comprehensive experiments on both the synthetic and proposed in-vivo datasets, along with user studies.
Extensive evaluations demonstrate that our novel paradigm outperforms previous rigid and non-rigid state-of-the-art registration methods, indicating its potential clinical applicability.
The main contributions are summarized below:

\begin{itemize}
	\item We present for the first time a self-supervised learning 3D-2D registration framework, named~\ourmodel, for preoperative-to-intraoperative landmark-free image fusion in laparoscopic liver surgery.
	\item We contribute a new in-vivo dataset for liver preoperative-to-intraoperative registration called \ourdata, which contains over 300 intraoperative frames with liver mask annotations and corresponding 3D point clouds.
	\item The structure-regularized shape adaptation (SRSA) strategy is proposed to model the global geometry similarity between preoperative and intraoperative models for coarse-to-fine deformable shape adaptation.
	%
	\item On both synthetic and in-vivo datasets, the proposed~\ourmodel~achieves superior performance over existing state-of-the-art rigid and non-rigid registration methods.
\end{itemize}

\section{related Work}
\subsection{3D Preoperative to 2D Intraoperative Registration}
Computer-assisted interventions for minimally invasive surgery have consistently aimed to achieve reliable registration between preoperative models and highly occluded intraoperative images.
Previous works~\cite{feuerstein2008intraoperative,zhang2019toward,espinel2024keyhole, lei2024epicardium} depended on relative static settings and requirements for additional devices or interaction inputs from surgeons, limiting their effectiveness in complex dynamic scenarios.
To eliminate the interaction during interventions, some methods~\cite{koo2022automatic,koo2017deformable,espinel2022using} focus on automatic registration by matching anatomical landmarks.
Although landmark-based pipelines have demonstrated promising registration performance, they struggle in complex scenarios with occlusions. 
To boost the deformable registration performance, several studies resort to the synthetic data simulated from the patient-specific organ by enforcing random forces~\cite{pfeiffer2020non, mhiri2024neural}.
However, these methods face limitations in the 3D-2D registration task, such as assuming the intraoperative 3D point cloud is known or overly relying on landmark detection.
In this regard, Collins \etal~\cite{collins2020augmented} proposed a markerless monocular AR system for uterus laparoscopic surgery, showing promising performance with few manual interactions. 
Wang \etal~\cite{wang2024video} presented a stereo-based AR navigation method for tissue deformation tracking in kidney surgery.
Zhang \etal~\cite{zhang2024point} proposed a keypoint-based correspondence learning method for liver 3D-3D registration.
Nevertheless, these methods rely on several idealized assumptions, \eg, the availability of intraoperative 3D meshes, relatively static organ deformations, and supplementary interactions from surgeons, which constrain their generalizability.
%
Besides, the inadequate exploration of intraoperative visual cues of the liver further leads to the degradation of registration accuracy.
In contrast, our work brings a landmark-free framework that addresses the 3D-2D registration via a self-supervised learning mechanism.

\vspace{-5pt}
\subsection{3D-3D Point Cloud Registration}
Point cloud registration is a widely researched topic in 3D vision, which aims to estimate the relative transformation between point cloud frames. 
It can be divided into two categories: rigid and non-rigid registration.
In the rigid setting, a variety of approaches have been proposed by using 1) traditional local descriptor matching~\cite{guo2013rotational} and iterative optimization~\cite{rusinkiewicz2019symmetric}; and 2) learning-based end-to-end regression~\cite{yew2022regtr} and correspondence-matching~\cite{huang2021predator,li2022lepard,qin2023geotransformer}.
In the non-rigid setting, a series of methods have emerged via various strategies, such as non-rigid iteration~\cite{li2008global}, deformable graph-based optimization~\cite{bozic2021neural}, and dense displacement or affine field-based estimation~\cite{li2022non}. 
A recent learning-based estimation method, Lepard~\cite{li2022lepard}, utilizes Transformer structure~\cite{vaswani2017attention} for robust point-to-point correspondence learning.
NDP~\cite{li2022non} brings a hierarchical deep affine transformation field via multiple MLPs for fast coarse-to-fine non-rigid registration. 
Inspired by above point cloud registration pipelines, we decouple the 3D-2D registration problem into 3D-3D rigid registration and non-rigid registration sub-problems, effectively alleviating the limitations of existing landmark-based methods.

\section{Dataset Acquisition and Processing}
\label{sec:Dataset}
\begin{figure}[t]
	\centering
	\includegraphics[width=0.49\textwidth]{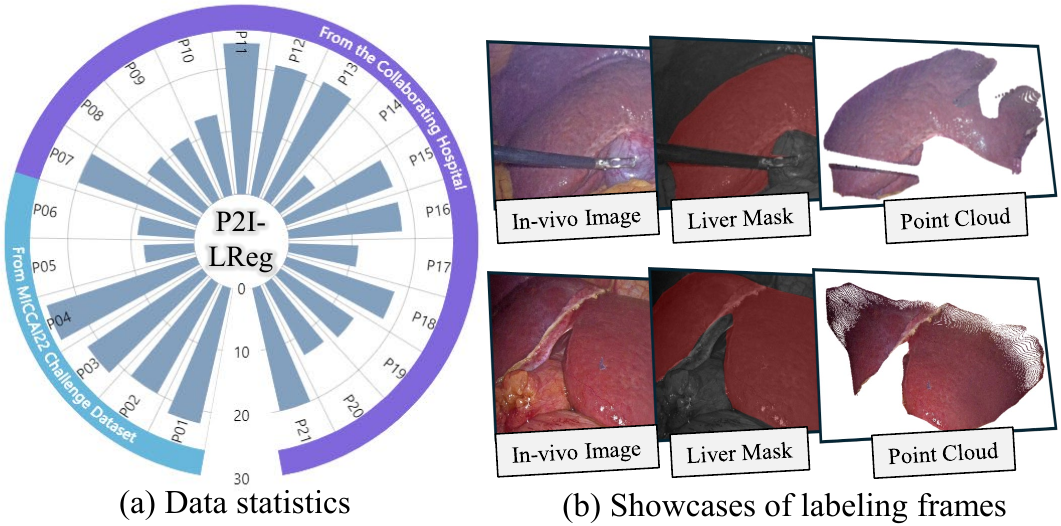}
	\caption{
		Illustration of our~\ourdata~dataset. 
	}
	\label{fig: data_generation}
\end{figure}

The available LLR 3D-2D registration datasets are extremely limited. The latest dataset~\cite{ALI2024103371} focuses on 2D and 3D liver landmark detection while lacking intraoperative 2D liver masks and 3D reconstructed liver models, as well as realistic 3D-2D registration ground truth. 
To bridge this gap, we develop a comprehensive in-vivo preoperative-to-intraoperative liver registration dataset named \ourdata. 
As shown in~\figref{fig: data_generation}(a), we collect 346 keyframes from liver resection videos of 21 patients with hepatocellular carcinoma (HCC). 
The annotations encompass the annotated liver segmentation masks and the reconstructed liver 3D point cloud for each keyframe, along with the preoperative 3D model segmented from CT images and the calibrated camera intrinsic matrix for each surgical video.
To enhance data diversity, we collect 15 surgical cases from partner hospitals, supplemented by six additional cases that are reprocessed from the MICCAI 2022 challenge~\cite{ALI2024103371}. 
For data processing, six senior surgeons are invited to select keyframes and annotate 2D liver masks, with four responsible for labeling and the other two for verification. 
The in-vivo samples with labels are shown in~\figref{fig: data_generation}(b), and the intrinsic parameters of the camera are calibrated by~\cite{zhang2000flexible}.
In \ourdata, we split twelve cases as the training set, four cases for validation, and five cases for testing. 

For training the rigid registration model, inspired by~\cite{pfeiffer2019generating}, we also construct a physics-based simulation environment in Blender~\footnote{https://www.blender.org/} for each patient-specific preoperative liver model, which is generally obtained weeks prior to surgery as part of routine preoperative imaging.
To simulate complex liver laparoscopic scenarios, we add scaled surgical instrument models together with other simulated tissues and organs such as fat, gallbladder, and ligaments.
For each rendered scene, the surgical instruments are consistently positioned within the view of the laparoscope and near the surface of the liver. 
Then, the laparoscope is randomly placed within a limited range targeting the liver.
In this setting, we generate a total of 2,500 RGB-D images for each patient.
	For all synthetic data, we randomly selected 20\% data for testing, 20\% for validation, and the remaining for registration network training.
All simulated data will be made publicly available to accelerate research in the liver registration community.

\begin{figure*}[th]
	\centering
	\includegraphics[width=0.99\linewidth]{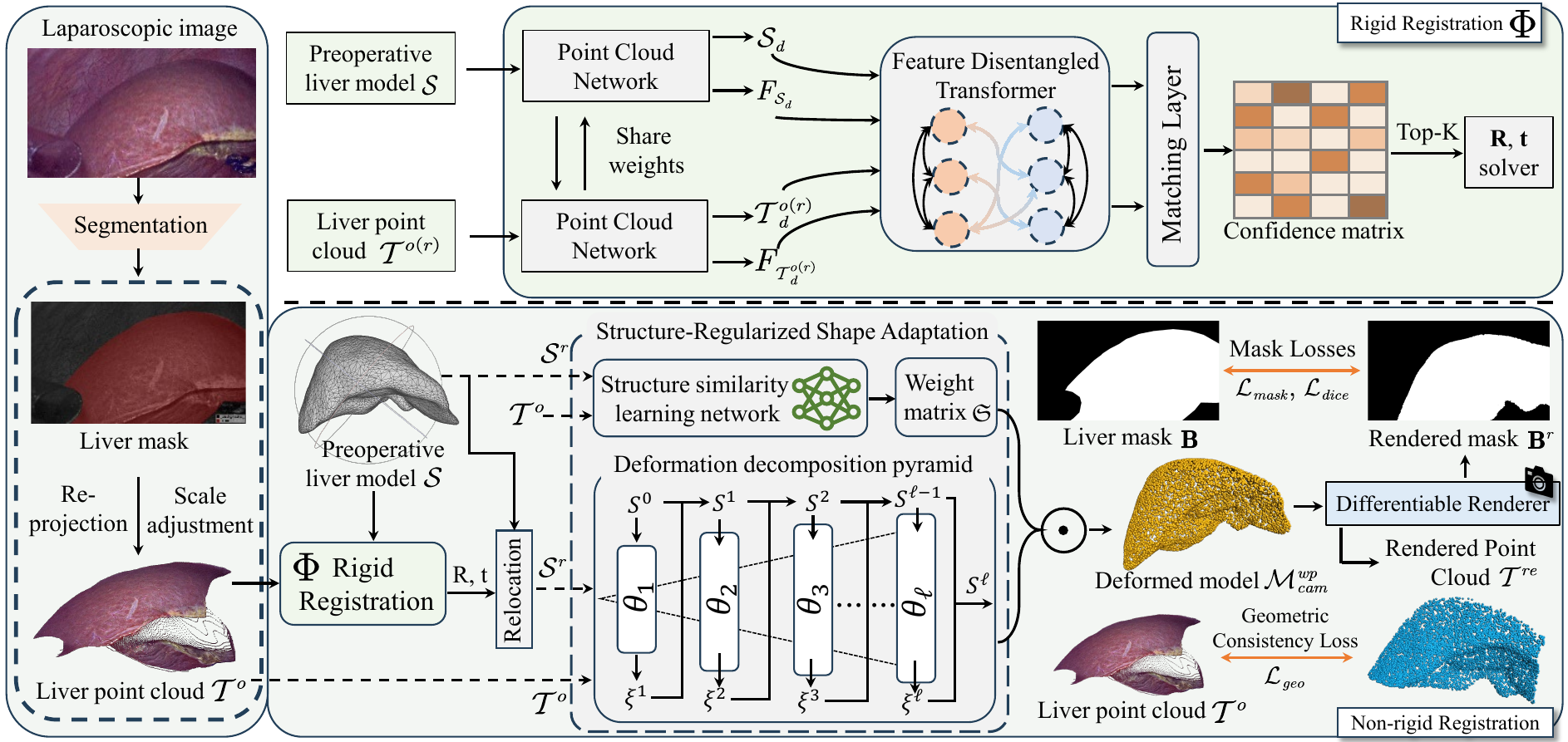}
	\caption{
		Overview of the proposed~\ourmodel. The synthetic data is firstly used to train the rigid registration network with a Feature Disentangled Transformer. For non-rigid registration, Structure-Regularized Shape Adaptation (SRSA) is designed to estimate the deformation field, comprising a deformation decomposition pyramid and a low-rank structure similarity learning network. During self-training on in-vivo data, a differentiable renderer is adopted to create the deformed, pose-dependent 2D liver mask, and 3D point cloud to compute the loss between rendered outputs and masks as well as the reprojected point cloud.
	}
	\label{fig: pipeline}
\end{figure*}

\vspace{-4pt}
\section{Methodology}
\subsection{Problem Statement}
We pioneered reforming the preoperative-to-intraoperative image fusion as a landmark-free 3D-3D point cloud registration task.
The challenge lies in recovering the partially visible liver 3D point cloud from monocular laparoscopic frames and registering the complete 3D liver shape onto partial liver point cloud in a landmark-free manner.
Let the point cloud $\mathcal{S}$$=$$\{\mathbf{s}_i$$\in$$\mathbb{R}^{3}\}^n_{i=1}$ denotes the vertices of the preoperative liver 3D model $\mathbf{M}$$=$$\{\mathcal{S},E,F\}$ segmented from CT images and $\mathcal{T}$$=$$\{\mathbf{t}_j$$\in$$\mathbb{R}^{3}\}^m_{j=1}$ represents visible liver surface points reprojected from the intraoperative 2D frame $\mathbf{I}$$\in$$\mathbb{R}^{H \times W \times 3}$, where $E,F$ denote the edges and faces respectively.
We aim to learn a warp function $\mathcal{W}: \mathbb{R}^3 \mapsto \mathbb{R}^3$ that aligns $\mathcal{S}$ to $\mathcal{T}$. 
In our setting, $\mathcal{W}$ can be parameterized by a $SE(3)$ transformation matrix~\footnote{Rigid transformations in 3D Euclidean distance space.} and a dense per-point warp field, \ie, rigid registration and non-rigid deformation. 
Using the camera intrinsic matrix $\mathcal{K}$$\in$$\mathbb{R}^{3\times3}$, the warped preoperative model $\mathbf{M}^{wp}$ is projected onto intraoperative images to yield 3D-2D registration results.

\subsection{Framework Overview}
%
As illustrated in~\figref{fig: pipeline}, we propose a self-supervised liver registration learning framework named~\ourmodel, which comprises two sub-networks: $SE(3)$ rigid registration network $\Phi$ and non-rigid deformation estimation network $\Psi$.
First, we utilize the patient-specific synthetic data (see~\secref{sec:Dataset}) to train $\Phi$. 
Taking the preoperative 3D model $\mathcal{S}$ from CT images and the simulated target rendered point cloud $\mathcal{T}^{r}$, which reflects intraoperative viewing conditions, as inputs, we leverage two weight-sharing geometric encoders to extract downsampled key points $\mathcal{S}_{d}$$\in$$\mathbb{R}^{\hat{n} \times 3}$, $\mathcal{T}^{r}_d$$\in$$\mathbb{R}^{\hat{m} \times 3}$ and corresponding multi-level geometric features $F_{\mathcal{S}_{d}}$$\in $$\mathbb{R}^{\hat{n} \times c}$, $F_{\mathcal{T}^{r}_d}$$\in$$\mathbb{R}^{\hat{m} \times c}$. 
Then, we feed them into the proposed feature disentangled transformer structure for global feature extraction and cross-geometrical cue aggregation.
After passing through the feature matching layer, the point-matching confidence matrix is obtained to produce the final correspondence between $\mathcal{S}_{d}$ and $\mathcal{T}^{r}_d$ by top $k$ confidence scores.
The rigid transformation matrices are solved by the SVD decomposition~\cite{besl1992method}.

After that, we freeze the pre-trained rigid registration network $\Phi$ and utilize it to infer $SE(3)$ transformation of $in$-$vivo$ data for subsequent liver deformable shape modeling. 
Given the observed intraoperative liver point cloud $\mathcal{T}^{o}$$\in$$\mathbb{R}^{m \times 3}$ (see~\secref{sec:pcd_recovery}) from the intraoperative image $\mathbf{I}$$\in$$\mathbb{R}^{H \times W \times 3}$ and preoperative model $\mathcal{S}$, we first re-locate $\mathcal{S}$ to the camera coordinate system $\mathcal{S}^r$ using the $SE(3)$ transformation predicted by $\Phi$. 
Then, the proposed structure-regularized shape adaptation module deforms $\mathcal{S}^r$ to $\mathcal{M}^{wp}_{cam}$, where the geometry of visible portions of the intraoperative liver is considered as a deformation geometric constraint for the preoperative model $\mathcal{S}^r$.
Finally, the deformed model $\mathcal{M}^{wp}_{cam}$ is projected onto $\mathbf{I}$ for preoperative-intraoperative registration (see~\secref{subsec:SRSA}).



\begin{figure*}[t]
	\centering
	\includegraphics[width=0.85\linewidth]{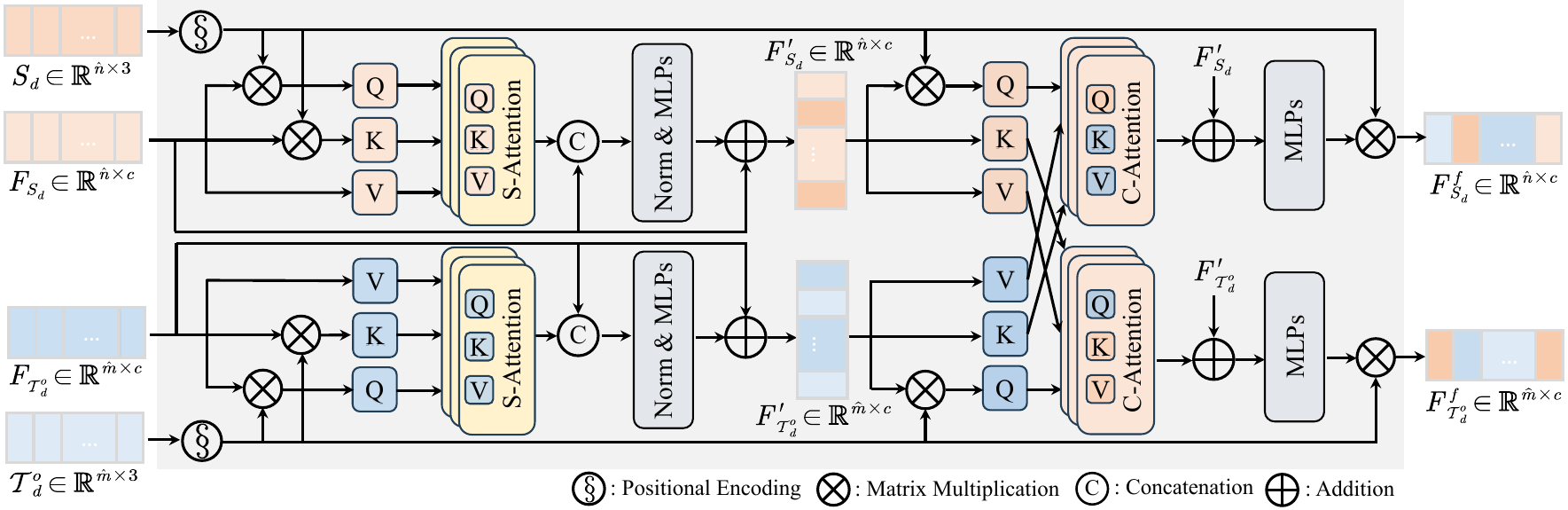}
	\caption{
		Illustration of our feature disentangled transformer structure.
	}
	\label{fig: transformer}
\end{figure*}

\subsection{Scale-Consistent 3D Liver Recovery}
\label{sec:pcd_recovery}
It is non-trivial to reconstruct an accurate liver 3D model from laparoscopy.
While laser scanning is ideal for acquiring liver geometry, it is impractical in surgical settings. 
While stereo matching has shown promise in depth estimation, monocular-based approaches remain more practical in real-world liver resection surgeries due to their lower cost and simplified hardware requirements~\cite{collins2020augmented}.
%

%
%
In this study, we focus on reconstructing a dense intraoperative liver point cloud model using monocular laparoscopic images, ensuring better scalability and clinical applicability.
Given the laparoscopic image $\mathbf{I}$, we adopt the depth-anything model~\cite{yang2024depth} to predict the relative depth map $\mathbf{D}$$\in$$\mathbb{R}^{H \times W}$, and then incorporating the annotated liver binary mask $\mathbf{B}$$\in$$\mathbb{R}^{H \times W}$, 
each 2D pixel is re-projected into the camera coordinate system to reconstruct 3D intraoperative liver point cloud $\mathcal{T}^{o}$:
\begin{equation}
	\mathcal{T}^{o} = \sigma\mathcal{K}^{-1}(\mathbf{D}\odot\mathbf{B})\mathbf{I},
\end{equation}
where $\sigma$ denotes the isotropic scale factor related to the liver size, $\mathcal{K}$ is the camera intrinsic matrix, and $\odot$ refers to the element-wise product.
Note that $\mathbf{D}$ only contains the relative depth value, indicating that the recovered point cloud has an unknown scale factor. 
Considering the consistent scale factor is indispensable for the 3D-3D registration pipeline \cite{li2022lepard} and the subsequent differentiable rendering, it is necessary to ensure the concordant scale factor between model $\mathcal{S}$ and $\mathcal{T}^{o}$.
To obtain the scale $\sigma$, we employ the principal component analysis (PCA)~\cite{wold1987principal} 
by taking the vertices $\mathcal{S}$ and $\mathcal{T}^{o}$ as inputs and compute the covariance matrix $C$ for them:
\begin{equation}
	C_{\mathcal{S},\mathcal{T}^{o}} = \frac{1}{\mathcal{N}_{\mathcal{S},\mathcal{T}^{o}}}\sum_{p \in \mathcal{S},\mathcal{T}^{o}} (p - \Bar{p})(p - \Bar{p})^T,
\end{equation}
where $\mathcal{N}_{\mathcal{S},\mathcal{T}^{o}}$, $\Bar{p}$ denote the number points and the center point of $\mathcal{S}$ and $\mathcal{T}^{o}$ respectively.
We calculate three eigenvalues and corresponding eigenvectors by decomposing matrix $C$. The eigenvector associated with the maximum eigenvalue serves as the principal axis, and the scale $\sigma_{\mathcal{S}}$ and $\sigma_{\mathcal{T}^{o}}$ can be obtained by projecting the points along the principal axis respectively. 
Taking the preoperative model as the reference, we can calculate the scale factor $\sigma$ by: $\sigma = \sigma_{\mathcal{S}} / \sigma_{\mathcal{T}^{o}}$.

\subsection{Preoperative-to-Intraoperative Image Fusion}

\subsubsection{Rigid Registration via Feature Disentangled Transformer}
Given the liver point clouds $\mathcal{S}$ and $\mathcal{T}^{o}$, we first utilize two weight-sharing point cloud networks (PCNs) $\Theta$ to extract geometric features. 
Concretely, we adopt the KPFPN backbone~\cite{thomas2019kpconv} to build $\Theta$, where the deformable KPConv layers are used for learning multi-level local geometry representation. 
The $\Theta$ outputs the downsampled points regarded as feature keypoints $\mathcal{S}_d$, $\mathcal{T}^{o}_{d}$ and the associated geometric features $F_{\mathcal{S}_d}$, $F_{\mathcal{T}^{o}_{d}}$.
We then feed them into our feature disentangled transformer for global context aggregation and cross-source interaction, enhancing the modeling capability of our model for liver geometric structures, as shown in~\figref{fig: transformer}.
Instead of only considering implicit semantic features~\cite{pei2024s}, we disentangle the point cloud into explicit geometric position embeddings and high-level geometric features. 
	This decoupled representation strategy has demonstrated its efficacy in registration tasks, especially in low-overlap cases \cite{qin2023geotransformer}.

Taking $\mathcal{S}_d$ as an example, we first adopt the Rotary positional encoding strategy~\cite{li2022lepard} for explicit relative geometric  embedding, formulated as $\S(\mathcal{S}_d) = [\mathbf{\Lambda}_1, \mathbf{\Lambda}_2, \cdots, \mathbf{\Lambda}_{c/6}]$, where $\S(\mathcal{S}_d)$ is a block diagonal matrix, $c$ is the number of channels of $F_{\mathcal{S}_d}$, and each block is a diagonal matrix with size 6$\times$6:
\begin{center}
	$
	\scalebox{0.74}{
		$
		\mathbf{\Lambda}_i=
			$\bigg[$
			\begin{pmatrix}
				\cos x\alpha_k  & -\sin x\alpha_k \\
				\sin x\alpha_k  & \cos x\alpha_k \\ 
			\end{pmatrix}
			$,$
			\begin{pmatrix}
				\cos y\alpha_k  & -\sin y\alpha_k  \\
				\sin y\alpha_k  & \cos y\alpha_k 
			\end{pmatrix}
			,
			\begin{pmatrix}
				\cos z\alpha_k  & -\sin z\alpha_k  \\
				\sin z\alpha_k  & \cos z\alpha_k 
			\end{pmatrix}
			$\bigg]$,
			$
		}    
		$
	\end{center}
	where $\alpha_k = 1/({10000^{6(k-1)/c}})$, \scalebox{0.9}{$k\in [1,2,..,c/6]$}.
	$\S(\mathcal{T}^{o}_{d})$ is also achieved by the same operation.
	Then, we compute the Query $Q$, Key $K$, and Value $V$ matrices for geometric features:
	\begin{equation}
		(Q, K) = \S(\mathcal{S}_d)(W_{q},W_{k})F_{\mathcal{S}_d}, \  V = W_{v}F_{\mathcal{S}_d},
	\end{equation}
	where $W_{q},W_{k}$ and $W_{v}$ $\in \mathbb{R}^{c\times c}$ are the projection matrices. 
	The enhanced feature $F^{\prime}_{\mathcal{S}_d} \in \mathbb{R}^{\hat{n} \times c}$ is calculated by:
	\begin{equation}
		F^{\prime}_{\mathcal{S}_d} = F_{\mathcal{S}_d} + LN(MLP(\mathcal{C}(F_{\mathcal{S}_d}, \varphi (Q, K, V)))),
		\label{equ:attention}
	\end{equation}
	where $LN(\cdot)$ and $\mathcal{C}(\cdot)$ are the layernorm and concatenation operation. 
	$\varphi (\cdot)$ is multi-head self-attention layers as in~\cite{vaswani2017attention}.
	The same operations are performed to produce $F^{\prime}_{\mathcal{T}^{o}_{d}}$$\in$$\mathbb{R}^{\hat{m} \times c}$.
	
	To enhance the interaction between cross-source features, we further feed $F^{\prime}_{\mathcal{S}_d}$ and $F^{\prime}_{\mathcal{T}^{o}_{d}}$ into a bidirectional cross-attention layer. 
	Similarly, the final deep cross-interacted feature $F_{S_d}^{f}$ and $F_{\mathcal{T}_{d}^{o}}^{f}$ are calculated by:
	\begin{equation}
		F^{f}_{\mathcal{S}_d} = \S(\mathcal{S}_d) \cdot MLP(F^{\prime}_{\mathcal{S}_d} + \varphi (Q^{{S}_d}, K^{\mathcal{T}_{d}^{o}}, V^{\mathcal{T}_{d}^{o}})),
	\end{equation}
	\begin{equation}
		F_{\mathcal{T}_{d}^{o}}^{f} = \S(\mathcal{T}_{d}^{o}) \cdot MLP(F^{\prime}_{\mathcal{T}^{o}_{d}} + \varphi (Q^{\mathcal{T}_{d}^{o}}, K^{\mathcal{S}_d}, V^{\mathcal{S}_d})),
	\end{equation}
	where the key difference is that the matrix $Q$ and $(K, V)$ stem from different source points corresponding to the direction of interaction, \ie, $\mathcal{S}_d$$\to$$\mathcal{T}_{d}^{o}$ or $\mathcal{T}_{d}^{o}$$\to$$\mathcal{S}_d$.
	Follow by~\cite{li2022lepard}, we then compute the confidence matrix $\mathbf{G}_{c} \in \mathbb{R}^{\hat{n} \times \hat{m}}$ for scoring the inlier point correspondences between $\mathcal{S}_d$ and $\mathcal{T}_{d}^{o}$:
	\begin{equation}
		\mathbf{G}_{c} = Softmax(\mathbf{\Gamma}(i,) \cdot Softmax(\mathbf{\Gamma}(,j)),
	\end{equation}
	\begin{equation}
		\mathbf{\Gamma}(i,j)=\langle W_{\mathcal{S}_d} F^{f}_{\mathcal{S}_d}, W_{\mathcal{T}_{d}^{o}} F_{\mathcal{T}_{d}^{o}}^{f} \rangle / \sqrt{c},
	\end{equation}
	where $W_{\mathcal{S}_d}$ and $W_{\mathcal{T}_{d}^{o}}$$\in$$\mathbb{R}^{c \times c}$ are learnable projection matrices. $\langle \cdot, \cdot \rangle$ denotes the inner product. The top $k$ inlier point correspondences are then selected based on a threshold $\eta_{H}$, \ie, $\hat{C} = \left\{(s_i, t_j)_r \mid s_i \in \mathcal{S}_d, t_j \in \mathcal{T}^o_{d} \right\}_{r=1}^{k}, \text{ s.t. } \mathbf{G}_{c}(i, j) > \eta_{H}$.
	
	Finally, given the correspondences $\hat{C}$ and the normalized confidence matrix $\widetilde{\mathbf{G}}_{c}$ of $\mathbf{G}_{c}$, we calculate the $SE(3)$ transformation: rotation $\mathbf{R} \in SO(3)$ and translation $\mathbf{t} \in \mathbb{R}^3$, via the SVD decomposition for the homography matrix $\mathbf{H}$$\in$$\mathbb{R}^3$:
	\begin{equation}
		\mathbf{H} = \sum\limits_{(s_{i},t_{j} \in \hat{C})}\widetilde{G}(i,j)_{c} s_{i} t_{j}^T = U\Sigma_{D}V^T
	\end{equation}
	Thus, the rotation $\mathbf{R}$$=$$VU^T$ and the translation $\mathbf{t}$$=$$(\sum(s_i) - \mathbf{R} \sum(t_j)) / \mathcal{N}_{\hat{C}}$, $\mathcal{N}_{\hat{C}}$ is the number of inlier matches in $\hat{C}$.
	

	\subsubsection{Non-Rigid Registration via Structure-Regularized Shape Adaptation}
	\label{subsec:SRSA}
	Due to the disruption by surgical tools and respiration, there is obvious shape deformation between $\mathcal{S}$ and $\mathcal{T}^o$. 
	To estimate the deformation of $\mathcal{S}$ and align it to $\mathcal{T}^o$, we first transform $\mathcal{S}$ to the camera coordinate system via the predicted rigid transformation to obtain an initialized alignment with $\mathcal{T}^o$, \ie, $\mathcal{S}^r = \mathbf{R}\mathcal{S} + \mathbf{t}$.
	Given $\mathcal{S}^r$ and $\mathcal{T}^o$, we then model the deformation field as a warp function with the mapping $\mathcal{W}_D: \mathbb{R}^3$$\to$$\mathbb{R}^3$. We learn $\mathcal{W}_D$ by a deformation decomposition strategy. 
	As shown in~\figref{fig: pipeline}, we first utilize the pyramid functions $\theta_{\ell}$$=$$(\Omega_{\ell}, \Xi_{\ell})$ as in~\cite{li2022non} to hierarchically represent the deformation field $\mathcal{W}_D$$=$$\{\theta_{\ell}\mid{\ell}=1,.., L\}$, where $\Omega_{\ell}$ and $\Xi_{\ell}$ denote the position encoding and MLP respectively. $L$ is the number of pyramid levels set to 10 in our experiments.
	For each level $\ell$, we have two mappings $\Omega_{\ell}: \mathbb{R}^3$$\to$$\mathbb{R}^6$ and $\Xi_{\ell}: \mathbb{R}^6$$\to$$\mathbb{R}^7$, \ie, $\Omega_{\ell}(s^{\ell-1}_{i}) =[\sin(2^{\ell+\ell_0} s^{\ell-1}_{i}), (\cos(2^{\ell+\ell_0} s^{\ell-1}_{i}))]$, $\Xi(\Omega_{\ell}(s^{\ell-1}_{i}))$$=$$(\xi^\ell_{i}, \vartheta^\ell_{i})$,
	where $s^{\ell-1}_{i} \in \mathcal{S}^r$ is the input point from the last level, $\ell_0$ controls the initial frequency at the first level. 
	$\xi^\ell_{i}$$\in$$\mathbb{R}^6$ and $\vartheta^\ell_{i}$$\in$$[0,1]$ denote the $SE(3)$ transformation and the confidence score respectively. The updated coordinate of $s_i$ at the $\ell$-th level can be calculated by:
	\begin{equation}
		s^{\ell}_{i} = s^{\ell-1}_{i} + \vartheta^\ell_{i} \cdot (\xi^\ell_{i} s^{\ell-1}_{i}).
	\end{equation}
	Accordingly, the transformation of each point is learned via a rigid-to-deformable sequence. 
	The overall deformation field $\mathcal{W}_D$ is obtained by combining these motions.
	However, direct application of this deformation strategy results in extreme model warping, i.e., $\mathcal{S}^r$ is fully warped to $\mathcal{T}^o$, which is impractical in clinical practice, as surgeons need to maintain a perception of the liver’s global anatomical context.
	Additionally, it is essential for the network not only to learn how to deform the preoperative model but also to know where to deform.
	To this end, we introduce a structure-regularized shape adaptation (SRSA) to constrain the estimated deformation field by extracting the robust similarity of two geometry features between intraoperative and preoperative models, injecting semantic features adaptively into $\mathcal{S}^r$ for subsequent adaptation.
	
	\begin{figure}[t]
		\centering
		\includegraphics[width=0.48\textwidth]{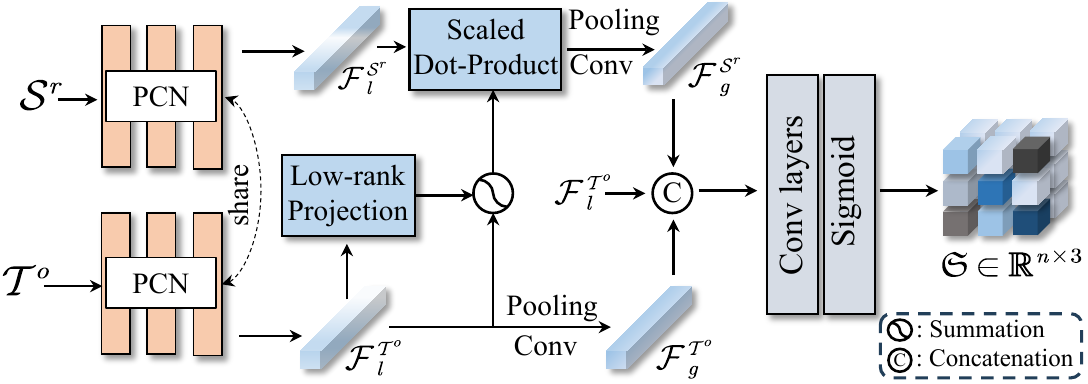}
		\caption{
			The low-rank structure similarity learning network. 
		}
		\vspace{-10pt}
		\label{fig: similarity}
	\end{figure}

	As illustrated in~\figref{fig: similarity}, we utilize two lightweight weights-shared PCN \cite{qi2017pointnet++} to extract local geometric features $\mathcal{F}^{\mathcal{S}^r}_{l}$, $\mathcal{F}^{\mathcal{T}^o}_{l}$$\in$$\mathbb{R}^{n(m) \times c_l}$. 
	For subsequent $\mathcal{S}^r$$\to$$\mathcal{T}^o$ structure adaptation, we adopt the inner product in the feature space to correlate the similar semantic features between $\mathcal{S}^r$ and $\mathcal{T}^o$.
	Considering the computation cost, we resort to the low-rank transformer structure~\cite{wang2020linformer} to yield the deformation weight matrix $\mathfrak{S}$:
	\begin{equation}
		\begin{aligned}
			&\mathfrak{S} = \phi_{s}\left(\phi_{c} \left ( \mathcal{C}(\mathcal{F}_{l}^{\mathcal{T}^o}, \phi_{cp}(\mathcal{F}_{l}^{\mathcal{T}^o}),\phi_{cp}(\varphi_{lr}(\mathcal{F}^{\mathcal{S}^r}_{l}, \mathcal{F}_{l}^{\mathcal{T}^o})) )\right) \right), \\
			&\varphi_{lr} = \delta(Q^s(\mathbb{E}_{K} K^s)^T / \sqrt{c_l}) \cdot (\mathbb{E}_{V} V^s),
		\end{aligned}
	\end{equation}
	where $\varphi_{lr}$ is the low-rank attention, $Q^s=W_q^s \mathcal{F}^{\mathcal{S}^r}_{l},K^s=W_k^s \mathcal{F}_{l}^{\mathcal{T}^o}$ and $V^s = W_v^s\mathcal{F}_{l}^{\mathcal{T}^o}$. 
	$\mathbb{E}_K$, $\mathbb{E}_V \in \mathbb{R}^{n\prime \times n}$ ($n\prime \ll n$) are two linear matrices that project $K^s$ and $V^s$ into low-dimension space.
	$\phi_s$, $\phi_c$, and $\phi_{cp}$ denote the Sigmoid function, convolution, and pooling layers.
	Subsequently, we incorporate $\mathfrak{S}$ into the finest level $\ell$ of the deformation pyramid, \ie, the level with the highest frequency. 
	The final warped preoperative liver model $\mathcal{M}_{cam}^{wp}$ of $S^r$ is generated by: 
	\begin{equation}
		\mathcal{M}_{cam}^{wp} = \mathfrak{S}\cdot\mathcal{W}_D(\mathcal{S}^r).
	\end{equation}
	In this way, our framework can effectively enhance the feature adaptability of the preoperative liver model, enabling adaptation to a variety of changing intraoperative liver instances.

	\emph{Preoperative-to-intraoperative image fusion:}
	we re-mesh the deformed model $\mathcal{M}_{cam}^{wp}$ by the pre-defined face indexes from $\mathbf{M}=\{\mathcal{S}, E, F\}$, noted as $\mathbf{M}_{cam}^{wp}$. Finally, the augmented image $\mathbf{I}^{aug}$ is obtained by projecting the meshed $\mathbf{M}_{cam}^{wp}$ onto the intraoperative laparoscopic image $\mathbf{I}$ incorporated with the camera intrinsic matrix $\mathcal{K}$, \ie, $\mathbf{I}^{aug} = \mathbf{I} + \mathcal{K} \mathbf{M}_{cam}^{wp}.$

	\begin{table*}[t]
		\setlength{\tabcolsep}{2.0pt}
		\scriptsize
		\centering
		\caption{
			Quantitative comparisons of rigid registration methods based on different correspondence samples and estimators.
		}
		\label{table:results-rigid}
		\resizebox{\linewidth}{!}{
			\begin{tabular}{l|ccccc|ccccc|ccccc|ccccc|ccccc}
				\toprule
				\# Metric & \multicolumn{5}{c|}{\emph{FMR} (\%) $\uparrow$} & \multicolumn{5}{c|}{\emph{IR} (\%) $\uparrow$} & \multicolumn{5}{c|}{\emph{RR} (\%) $\uparrow$} & \multicolumn{5}{c|}{\emph{RRE} ($^\circ$) $\downarrow$} & \multicolumn{5}{c}{\emph{RTE} (mm) $\downarrow$}\\
				\# Correspondence Samples & 2000 & 1500 & 1000 & 500 & 250 & 2000 & 1500 & 1000 & 500 & 250 & 2000 & 1500 & 1000 & 500 & 250 & 2000 & 1500 & 1000 & 500 & 250 & 2000 & 1500 & 1000 & 500 & 250\\
				\midrule
				\midrule
				Estimator & \multicolumn{25}{c}{RANSAC-\emph{50k}~\cite{fischler1981random}} \\
				\midrule
				\midrule
				Predator (CVPR'21)~\cite{huang2021predator} & 95.5 & 95.4 & 95.3 & 95.4 & 95.3 & 78.6 & 75.3 & 73.9 & 70.3 & 66.1 & 89.2 & 89.3 & 89.0 & 88.2 & 86.7 & 0.287 & 0.286 & 0.288 & 0.290 & 0.296 & 2.14 & 2.14 & 2.16 & 2.17 & 2.21 \\
				GeoTrans (TPAMI'23)~\cite{qin2023geotransformer} & 96.7 & \textbf{96.9} & \textbf{96.9} & 96.7 & 96.5 & 82.7 & 88.9 & 94.7 & 96.2 & 96.8 & 94.5 & 94.2 & 93.5 & 93.9 & 94.6 & 0.219 & 0.220 & 0.224 & 0.221 & 0.219 & 1.38 & 1.39 & 1.44 & 1.41 & 1.38 \\ 
				\rowcolor{gray!20} Self-P2I (ours) & \textbf{96.9} & \textbf{96.9} & 96.8 & \textbf{96.8} & \textbf{96.8} & \textbf{87.8} & \textbf{92.4} & \textbf{95.0} & \textbf{97.6} & \textbf{98.1} & \textbf{98.8} & \textbf{98.9} & \textbf{99.0} & \textbf{98.9} & \textbf{98.8} & \textbf{0.195} & \textbf{0.193} & \textbf{0.190} & \textbf{0.189} & \textbf{0.194} & \textbf{1.23} & \textbf{1.22} & \textbf{1.21} & \textbf{1.21} & \textbf{1.24} \\
				\midrule
				\midrule
				Estimator & \multicolumn{25}{c}{LGR~\cite{qin2023geotransformer}} \\
				\midrule
				\midrule
				Predator (CVPR'21)~\cite{huang2021predator} & 95.3 & 95.3 & 95.2 & 95.4 & 95.2 & 75.6 & 74.1 & 70.3 & 67.4 & 64.9 & 88.5 & 88.4 & 87.6 & 86.9 & 85.2 & 0.287 & 0.289 & 0.294 & 0.298 & 0.307 & 2.22 & 2.23 & 2.26 & 2.30 & 2.37 \\
				GeoTrans (TPAMI'23)~\cite{qin2023geotransformer} & \textbf{96.7} & \textbf{96.7} & \textbf{96.7} & \textbf{96.5} & \textbf{96.6} & 82.3 & 87.9 & 93.8 & 96.1 & 96.9 & 94.3 & 93.4 & 92.9 & 93.0 & 93.2 & 0.219 & 0.227 & 0.228 & 0.227 & 0.227 & 1.40 & 1.45 & 1.51 & 1.49 & 1.47 \\ 
				\rowcolor{gray!20} Self-P2I (ours) & 96.4 & 96.3 & 96.1 & 96.3 & 96.4 & \textbf{84.7} & \textbf{91.7} & \textbf{94.2} & \textbf{96.6} & \textbf{97.8} & \textbf{96.2} & \textbf{95.7} & \textbf{96.2} & \textbf{96.2} & \textbf{96.4} & \textbf{0.209} & \textbf{0.207} & \textbf{0.207} & \textbf{0.201} & \textbf{0.203} & \textbf{1.29} & \textbf{1.28} & \textbf{1.28} & \textbf{1.27} & \textbf{1.31}  \\
				\midrule
				\midrule
				Estimator & \multicolumn{25}{c}{weighted SVD~\cite{besl1992method}} \\
				\midrule
				\midrule
				Predator (CVPR'21)~\cite{huang2021predator} & 95.4 & 95.4 & 95.3 & 95.3 & 95.5 & 78.4 & 75.2 & 73.1 & 69.7 & 65.5 & 86.9 & 86.3 & 85.6 & 84.9 & 84.7 & 0.299 & 0.302 & 0.307 & 0.314 & 0.314 & 2.29 & 2.31 & 2.35 & 2.39 & 2.39 \\
				GeoTrans (TPAMI'23)~\cite{qin2023geotransformer} & \textbf{96.9} & \textbf{96.9} & 96.8 & \textbf{96.9} & \textbf{96.7} & 82.9 & 88.6 & 92.8 & 96.3 & 97.1 & 92.7 & 91.4 & 91.1 & 91.8 & 91.6 & 0.251 & 0.255 & 0.257 & 0.254 & 0.254 & 1.54 & 1.63 & 1.63 & 1.59 & 1.61\\ 
				\rowcolor{gray!20} Self-P2I (ours) & \textbf{96.9} & \textbf{96.9} & \textbf{96.9} & 96.8 & \textbf{96.7} & \textbf{86.9} & \textbf{92.0} & \textbf{93.6} & \textbf{96.8} & \textbf{97.8} & \textbf{94.8} & \textbf{93.7} & \textbf{94.0} & \textbf{94.3} & \textbf{95.1} & \textbf{0.217} & \textbf{0.215} & \textbf{0.212} & \textbf{0.211} & \textbf{0.216} & \textbf{1.35} & \textbf{1.32} & \textbf{1.34} & \textbf{1.32} & \textbf{1.38} \\
				\bottomrule
			\end{tabular}
		}
	\end{table*}
	
	
	\subsection{Self-Supervised Learning and Loss Function}
	We train our framework in a self-supervised learning manner via intraoperative 2D liver mask $\mathbf{B}$ and re-projecting 3D point cloud $\mathcal{T}^o$. 
	To eliminate the partial-to-complete inconsistency between ($\mathbf{B}, \mathcal{T}^o$) and $\mathcal{M}_{cam}^{wp}$, we utilize a differentiable renderer~\cite{ravi2020pytorch3d} for $\mathcal{M}_{cam}^{wp}$ to render the deformed pose-dependent depth map $\mathbf{D}^r$ and shape mask $\mathbf{B}^r$. 
	We lift the valid depth value back to point cloud to attain visible liver surface points $\mathcal{T}^{re}$.
	In short, if the predicted pose $[\mathbf{R}, \mathbf{t}]$ and deformation field $[\mathfrak{S}\mathcal{W}_D]$ are accurate, $\mathcal{T}^{re}$, $\mathbf{B}^r$ would be well aligned with the intraoperative liver point cloud $\mathcal{T}^o$ in 3D space and shape mask $\mathbf{B}$ in 2D space. 
	Overall, the total loss function is:
	\begin{equation}
		\mathcal{L}=  \lambda_{dice}\mathcal{L}_{dice} + \lambda_{mask}\mathcal{L}_{mask} + \lambda_{geo}\mathcal{L}_{geo},
	\end{equation}
	where $\mathcal{L}_{dice}$ and $\mathcal{L}_{mask}$ are used to measure the silhouette similarity between rendered and ground truth masks. 
	$\lambda_{\{dice, mask, geo\}}$ denote the balancing weights.
	Here, we use the Dice loss as $\mathcal{L}_{dice}$.
	$\mathcal{L}_{mask}$ is formulated by the binary cross-entropy loss.
	Besides, $\mathcal{L}_{geo}$ measures the geometric consistency between the rendered and intraoperative liver point clouds, which is defined as
	\begin{equation}
		\mathcal{L}_{geo} = \frac{1}{N_1} \sum_{x \in \mathcal{T}^{re}} \min_{y \in \mathcal{T}^o}||x - y||^2_2 + \frac{1}{N_2} \sum_{y \in \mathcal{T}^o} \min_{x \in \mathcal{T}^{re}}||y - x||^2_2 ,
		\label{equ:cd}
	\end{equation}
	where $N_1$, $N_2$ denote the number of points for $\mathcal{T}^{re}$ and $\mathcal{T}^o$. 
	
	For training the rigid registration network $\Phi$, we use the loss functions $\mathcal{L}_{corr}$ and $\mathcal{L}_{tran}$ in~\cite{li2022lepard} for inlier correspondences confidence matrix and final transformation supervision. 
	$\mathcal{L}_{corr}$ is defined by the standard focal loss to minimize the distance between the predicted confidence matrix $\mathbf{G}_{c}$ and the ground truth. 
	$\mathcal{L}_{tran}$ is adopted to minimize the reprojection error over the predicted pose $[\mathbf{R}, \mathbf{t}]$ and the ground truth. 
	For training the deformation pyramid $\mathcal{W}_D$, we use Eq.~(\ref{equ:cd}) at each level $\ell$ to minimize the chamfer distance between $\mathcal{M}_{cam}^{wp}$ and $\mathcal{T}^o$.

\section{Experiments and Results}
\subsection{Experiment Setting and Implementation Details}
Our framework is implemented with PyTorch and trained on two NVIDIA RTX 3090 GPUs.
We fixed the size of the input point clouds to 8192 points and downsampled it with a voxel size of $0.001m$ before feeding into the KPFPN backbone~\cite{thomas2019kpconv}.
During the training phase, we use the SGD optimizer to train our rigid registration network $\Phi$ for 120 epochs, and then fix the parameters and use the Adam optimizer to train the overall framework for 80 epochs.
The initial learning rates are all set to $1\mathrm{e}^{-4}$ and the batch size is 2. 
For training $\Phi$, we set all loss weights to 1 in line with~\cite{li2022lepard}.
For overall loss function $\mathcal{L}$, we set $\lambda_{dice}$, $\lambda_{mask}$, and $\lambda_{geo}$ to 1. The threshold $\eta_{H}$ is set to 0.15 in our experiments.

We evaluate the effectiveness of our proposed approach on both synthetic and in-vivo datasets.
Due to differences in task configurations, it is inappropriate to compare our method with previous landmark-based approaches, which measure similarity between landmarks~\cite{ALI2024103371}, while landmarks are not used and annotated in our dataset. Therefore, given that our framework is inspired by point cloud registration, we propose to compare it with state-of-the-art point cloud registration methods applicable to this task to validate its efficacy.
All competitors are retrained from scratch for a fair comparison.

For in-vivo data evaluation, the results were obtained using five-fold cross-validation. For each method, we conducted five rounds of training with a different 12/4/5 split for training, validation, and testing at the patient level. We report the mean performance across these five rounds.
Additionally, we randomly selected a test set from one of the five folds for further quantitative and qualitative analysis, without altering the original data partition.

\begin{figure}[t]
	\centering
	\includegraphics[width=0.48\textwidth]{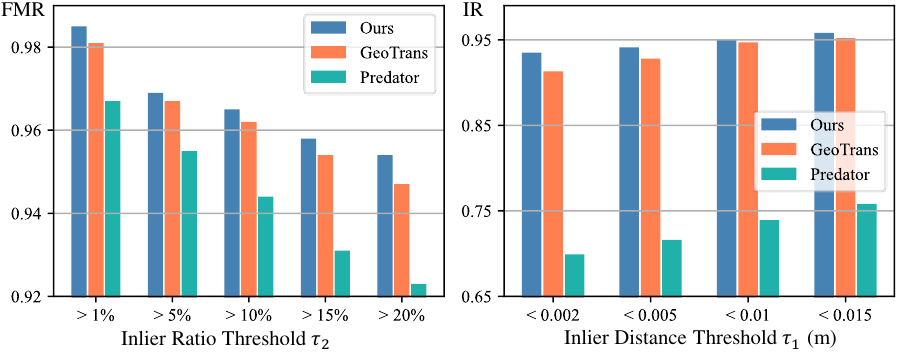}
	\caption{
		Performance of: FMR with the inlier ratio threshold ($\tau_2$), and IR about the Inlier distance Threshold ($\tau_1$).
	}
	\label{fig: FMR-IR-T}
\end{figure}

\subsection{Evaluation Metrics}
We adopt standard evaluation metrics commonly used in point cloud registration and preoperative-to-intraoperative registration tasks.
For rigid registration validation, we use five metrics following \cite{huang2021predator,qin2023geotransformer,li2022lepard}:

($\romannumeral1$) $Registration\ Recall$ (RR), the fraction of point cloud pairs with transformation errors below a threshold $\rho$, indicating overall registration performance.
($\romannumeral2$) $Inlier\ Ratio$ (IR), the fraction of speculative matches with distances under a threshold $\tau_1$ using the ground-truth transformation, measuring the correctness of putative matches.
($\romannumeral3$) $Feature\ Matching\ Recall$ (FMR), the fraction of IR$>$$\tau_2$=5$\%$ in all point cloud pairs, assessing the likelihood of accurately recovering transformation parameters from inlier correspondences.
($\romannumeral4$) $Relative\ Rotation\ Error$ (RRE), the geodesic distance between predicted and ground-truth rotation matrices.
($\romannumeral5$) $Relative\ Translation\ Error$ (RTE) measures the Euclidean distance between putative and ground-truth translation vectors.
%
For non-rigid registration validation, we refer to metrics of the P2ILF challenge~\cite{ALI2024103371}, including the ($\romannumeral6$) \emph{Dice coefficient} and ($\romannumeral7$) \emph{Chamfer distance} (CD) metrics. 
The former measures the similarity between annotated liver masks and projected deformed models in 2D space; the latter measures the similarity between intraoperative liver point clouds and deformed preoperative models in 3D space.
Furthermore, we conduct the ($\romannumeral8$) $User\ Study$ by inviting expert surgeons to evaluate the registration results using rating questionnaires.

\begin{figure}[t]
	\centering
	\includegraphics[width=0.5\textwidth]{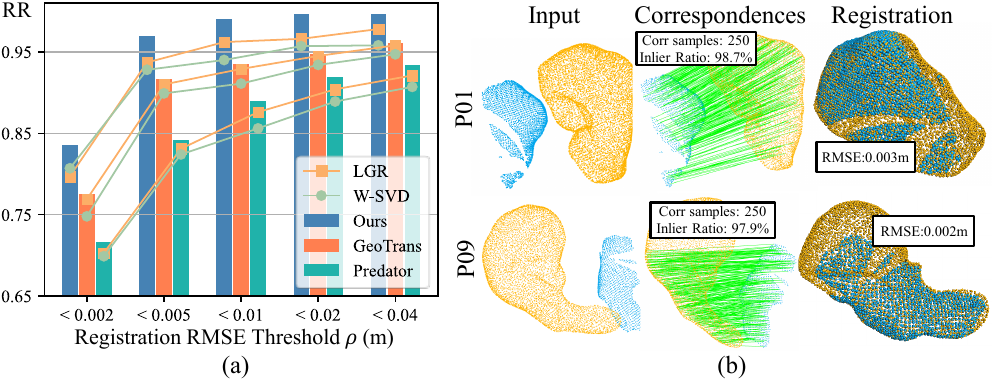}
	\caption{
		(a) Performance of RR in relation to the threshold $\rho$. (b) Qualitative rigid registration results. The preoperative, intraoperative models, and correspondences are shown in \textcolor{yellow}{yellow}, \textcolor{blue}{blue}, and \textcolor{green}{green}, respectively. The lower RMSE, the better.
	}
	\label{fig: RR-T}
\end{figure}

\subsection{Comparison with State-of-the-Art Methods}\label{comp_sota}

\subsubsection{Rigid Registration Evaluation}
We first evaluate the rigid registration performance on the synthetic dataset by comparing~\ourmodel~with the cutting-edge point cloud registration methods: Predator~\cite{huang2021predator} and GeoTrans~\cite{qin2023geotransformer}. 
Following~\cite{qin2023geotransformer,huang2021predator}, we employ the top-k sampling scheme to report quantitative results with varying numbers of correspondences in~\tabref{table:results-rigid}.
RANSAC-$\emph{50k}$ (50K iterations)~\cite{fischler1981random} is adopted as the default estimator to compute the rigid transformation.
%


\noindent$\romannumeral1$) \textbf{\emph{Matching learning results.}}
The 2$^{nd}$-3$^{rd}$ columns in~\tabref{table:results-rigid} showcase the correspondence learning quality results.
For the FMR metric, our method outperforms Predator by at least 1.4\% at all sample levels, demonstrating the effectiveness of our model on feature representation learning.
For the IR metrics, our method also significantly surpasses other methods, with improvements of 9\%$\sim$32\% and 0.3\%$\sim$5.1\% over Predator and GeoTrans, respectively.
It reveals that~\ourmodel~can extract reliable point correspondences.
To verify the robustness of estimated correspondences, we also follow ~\cite{qin2023geotransformer,huang2021predator} to show the relationship between FMR and IR with varying threshold levels $\tau_2$ and $\tau_1$. As shown in~\figref{fig: FMR-IR-T}, our method maintains robust performance at different threshold levels.
\noindent$\romannumeral2$) \textbf{\emph{Registration results.}}
Columns 4$^{th}$-6$^{th}$ of~\tabref{table:results-rigid} present the registration comparative results. 
On the important RR metric, our method surpasses other methods by a large margin. 
For RRE and RTE metrics, our model also achieves better scores with minimal rotation and translation errors.
\figref{fig: RR-T}(a) shows the registration results at different threshold levels. 
Our model improves significantly, especially at the 5 mm threshold,  maintaining a registration recall above 95\%.
\figref{fig: RR-T}(b) exhibits the qualitative registration results on the varying liver anatomical structures. 
\ourmodel~provides an accurate partial-to-complete liver registration by extracting reliable matches.

\begin{table}[!t]
	\setlength{\tabcolsep}{1pt}
	\centering
	\caption{Quantitative comparison with five-fold cross-validation on \ourdata~dataset. 
	}
	\label{tab:non-rigid}
	\setlength{\tabcolsep}{8pt}
	\resizebox{\linewidth}{!}
	{ 
		\begin{tabular}{l|cc}
			\hline
			& \multicolumn{2}{c}{MEAN}
			\cr\hline
			Models  & DICE (\%) $\uparrow$   & CD (mm) $\downarrow$     
			\cr\hline
			
			Sinkhorn (AISTATS'19)~\cite{feydy2019interpolating} & 42.01\ci{5.26} & 5.85\ci{1.32} \\
			Nerfies (ICCV'21)~\cite{park2021nerfies} & 64.07\ci{4.91} & 3.90\ci{0.88} \\
			NSFP (NeurIPS'21)~\cite{li2021neural} & 62.03\ci{4.97} & 3.92\ci{0.84} \\
			NDP (NeurIPS'22)~\cite{li2022non} & 64.77\ci{5.68} & 3.75\ci{0.87} \\
			GeoTrans (TPAMI'23)~\cite{qin2023geotransformer} & 71.16\ci{6.46} & 3.43\ci{0.96} \\
			DPF (ICCV'23)~\cite{Prokudin_2023_ICCV} & 72.19\ci{5.97} & 3.31\ci{1.18} \\
			PointSetReg (CVPR'24)~\cite{zhao2024clustereg} & 75.24\ci{5.92} & 3.20\ci{1.03} \\
			
			\ourcell \textbf{\ourmodel~(Ours)} & \ourcell \textbf{78.89}\ci{6.76} & \ourcell \textbf{2.97}\ci{1.06}\\
			
			\hline

		\end{tabular}
	}
\end{table}

\begin{figure}[t]
	\centering
	\includegraphics[width=0.49\textwidth]{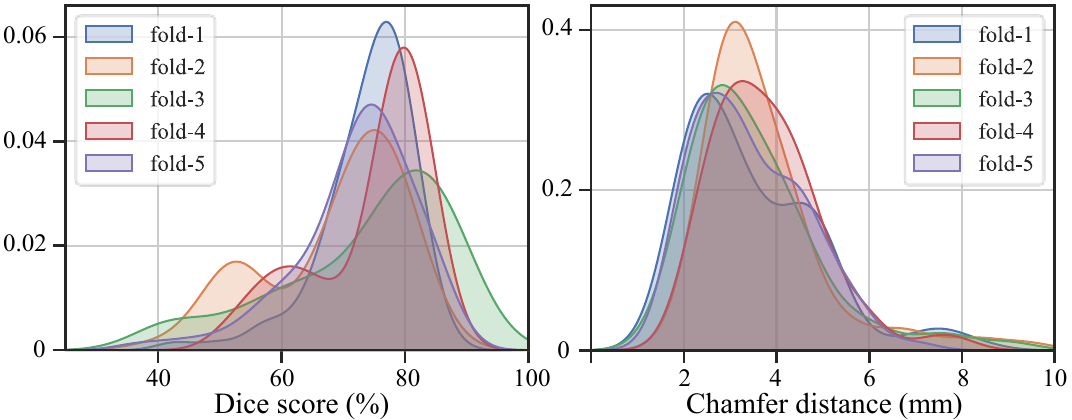}
	\caption{
		Density distribution of Dice score and Chamfer distance across all five folds. Y axis: Density.
	}
	\label{fig: density_distribution}
\end{figure}

\begin{figure*}[t]
	\centering
	\includegraphics[width=\linewidth]{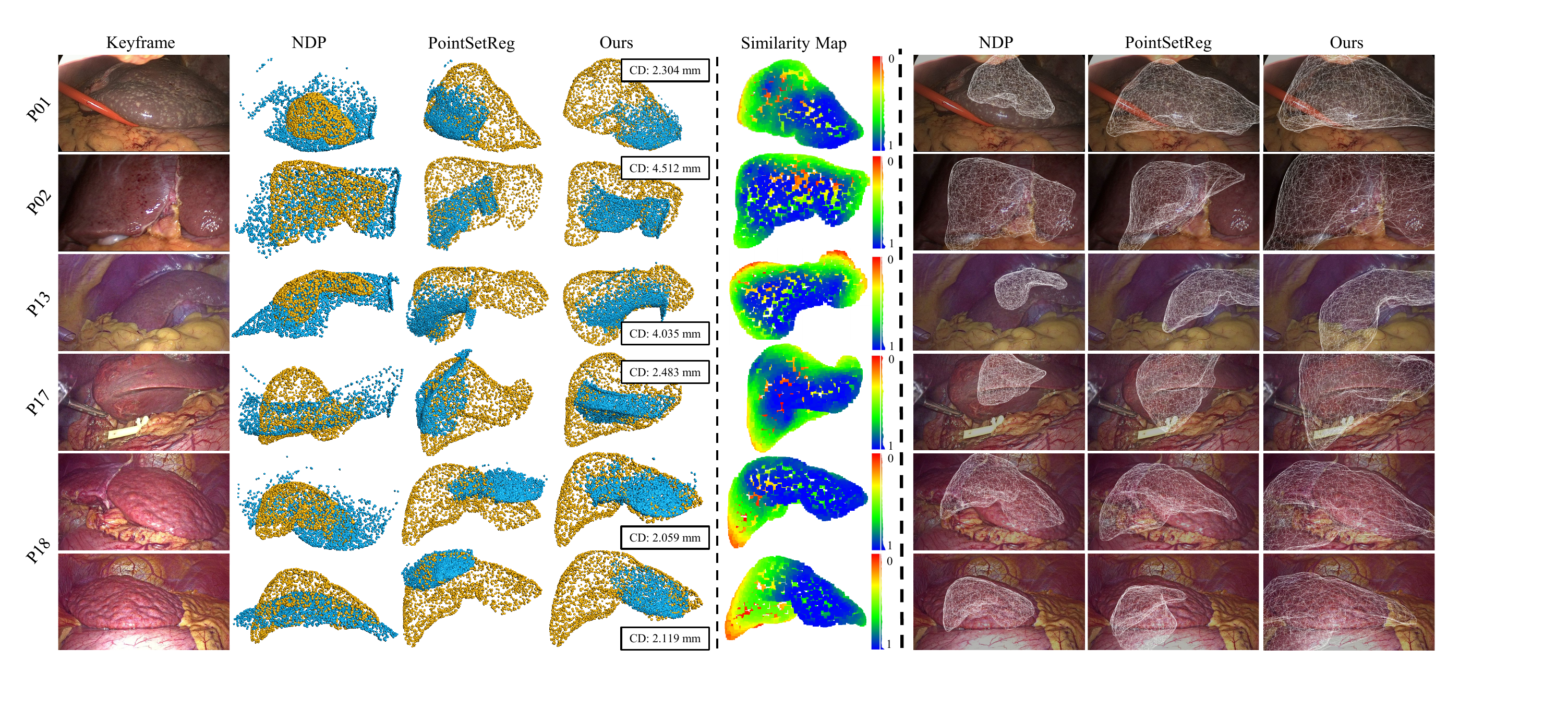}
	\caption{
		Qualitative results from the randomly selected one-fold test set.  The column(s) of 1$^{st}$ is the keyframe from the surgical video; 2$^{nd}$-4$^{th}$ and 6$^{th}$-8$^{th}$ are the comparison results of 3D-3D and 3D-2D registration respectively; 5$^{th}$ is the similarity map learned by our proposed SRSA (\secref{subsec:SRSA}). Preoperative and intraoperative models are shown in \textcolor{yellow}{yellow} and \textcolor{blue}{blue} (2$^{nd}$-4$^{th}$).
	}
	\label{fig: main_vis}
\end{figure*}

\noindent$\romannumeral3$) \textbf{\emph{Results with varying estimators.}}
Drawing on the strategy in~\cite{qin2023geotransformer}, we also compare different pose estimators (\ie, LGR~\cite{qin2023geotransformer} and weighted SVD~\cite{besl1992method}) to eliminate the effect of the pose solver.
As shown in~\tabref{table:results-rigid}, without using RANSAC for outlier filtering, our model achieves superior performance with both the local-to-global (LGR) estimator~\cite{qin2023geotransformer} and the Weighted SVD estimator~\cite{besl1992method}. 
Further, \figref{fig: RR-T}(a) (line chart) report the registration recall against varying RMSE thresholds $\rho$. We can see that~\ourmodel~reaches optimal registration recall at each threshold. When $\rho$$<$$0.005m$, our method still maintains RR over 92\%, showcasing its effectiveness and robustness.

\begin{figure}[t]
	\centering
	\includegraphics[width=0.46\textwidth]{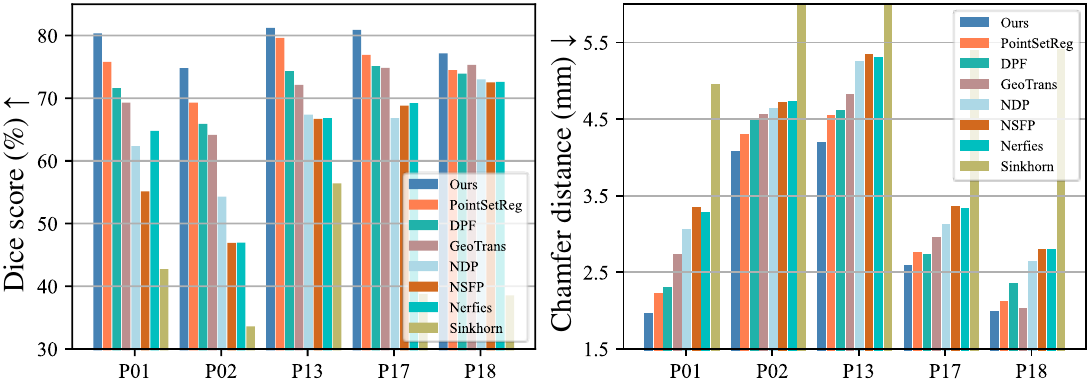}
	\caption{
		Results on the randomly selected one-fold test set. 
	}
	\label{fig: one-fold_performance}
\end{figure}

\subsubsection{Non-rigid Registration Evaluation}

We compare our model with well-performed point cloud deformation methods~\cite{feydy2019interpolating,qin2023geotransformer,zhao2024clustereg,park2021nerfies,li2021neural,li2022non,Prokudin_2023_ICCV} in 3D and 2D space to evaluate the non-rigid registration performance on our \ourdata~datasets.

\noindent$\romannumeral1$) \textbf{\emph{Registration in 3D space.}}
\tabref{tab:non-rigid} shows the comparisons for 3D-3D registration. Our method obtains better results on the CD metric, proving its effectiveness for registration in 3D space. 
\figref{fig: one-fold_performance} also displays the results of the randomly selected one-fold test set. Compared to baseline (NDP~\cite{li2022non}), our method reduces the CD value in all test cases by 0.54 mm$\sim$1.09 mm, which benefits from our structure-regularized shape adaptation design. 
Qualitative results in~\figref{fig: main_vis} show that our method provides accurate deformation by measuring structural similarity (5$^{th}$ column) to achieve precise registration, whereas NDP and PSR produce unreasonable shape warping leading to incorrect registration results (2$^{nd}$-4$^{th}$ columns). 

\noindent$\romannumeral2$) \textbf{\emph{Registration in 2D space.}}
\tabref{tab:non-rigid} also reports the 3D-2D registration comparison results.
\ourmodel~shows 14.12\% and 3.65\% increase on average Dice, with better performance in individual cases across all patient samples (\figref{fig: one-fold_performance}).
The last three columns of~\figref{fig: main_vis} also exhibit comparisons of the 3D preoperative model to 2D intraoperative image registration.
We can observe that competitors may fail to estimate the pose parameters resulting in distortion of the preoperative model to a sub-optimal state, whereas our method obtains a more accurate 3D-2D registration with the proposed SRSA strategy.

Fig.~\ref{fig: density_distribution} further presents the density distribution of the registration results on the test sets of all folds. Each curve aggregates results over all test cases in a given fold. The consistent trends across folds demonstrate the robustness and generalizability of our method.

\subsubsection{User Study}

\begin{figure}[t]
	\centering
	\includegraphics[width=0.46\textwidth]{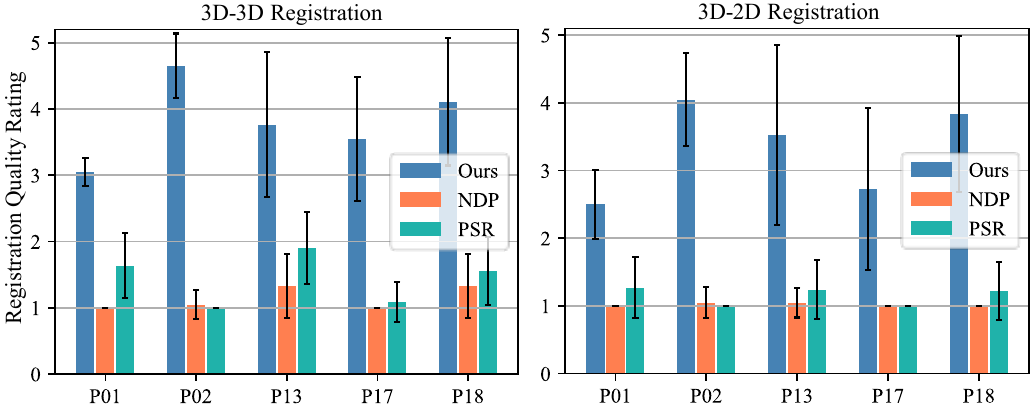}
	\caption{
		User studies of the randomly selected one-fold test set from the proposed \ourdata~dataset. PSR: PointSetReg~\cite{zhao2024clustereg}.
	}
	\label{fig: user study}
\end{figure}

We invite six expert surgeons to rate each registration result for randomly selected one-fold test cases to compare our method more comprehensively with competitors. 
The user study includes two scoring questions: (1) How is the 3D-3D registration quality? (2) How is the 3D-2D registration quality? 
The rating scores range from 1 to 5, with 5 indicating the best. 
From the results in~\figref{fig: user study}, we can draw two conclusions: 
$\romannumeral1$) The registration performance of the proposed~\ourmodel~is superior than existing methods. 
$\romannumeral2$) The surgeons are satisfied with our registration results in most cases. 
The findings prove the superiority of our method and the potential for clinical application in liver surgery.

\begin{table}[t]
	\setlength{\tabcolsep}{1.1pt}
	\scriptsize
	\centering
	\caption{
		Effect of varying training strategies for network $\Phi$.
	}
	\label{table:ab_rigid_training_strategy}
	\resizebox{0.95\columnwidth}{!}{  
		\begin{tabular}{l|cc|cc|c}
			\hline
			\multirow{2}{*}{Strategies} & \multicolumn{2}{c|}{\emph{Rigid}} & \multicolumn{2}{c|}{Non-rigid} & \multicolumn{1}{c}{Training Time} \\
			& \emph{IR} (\%) $\uparrow$ & \emph{RR} (\%) $\uparrow$ & \emph{Dice} (\%) $\uparrow$ & \emph{CD} (mm) $\downarrow$ & (hours) $\downarrow$ \\
			\hline
			$\Phi$ \   \includegraphics[height=1em]{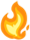} & 94.6 & 98.4 & 72.46$\pm$7.49 & \textbf{2.71$\pm$1.28} & $\sim$ 14.0 \\
			\rowcolor{gray!20} $\Phi$ \   \includegraphics[height=1em]{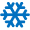} & \textbf{95.0} & \textbf{99.0} & \textbf{78.89$\pm$6.76} & 2.97$\pm$1.06 &  $\sim$ \textbf{9.3}  \\
			\hline
		\end{tabular}
	}
\end{table}

\begin{table}[t]
	\setlength{\tabcolsep}{1.1pt}
	\scriptsize
	\centering
	\caption{
		Effect of varying loss weights for rigid registration.
	}
	\label{table:ab_loss_weight_rigid}
	\resizebox{0.95\columnwidth}{!}{  
		\begin{tabular}{c|c|c|ccc|c|c}
			\hline
			\multirow{2}{*}{$\lambda_{tran}$} & \multicolumn{1}{c|}{\emph{FMR} (\%) $\uparrow$} & \multicolumn{1}{c|}{\emph{IR} (\%) $\uparrow$} & \multicolumn{3}{c|}{\emph{RR} (\%) $\uparrow$ ($\rho$)} & \multicolumn{1}{c|}{\emph{RRE} $\downarrow$} & \multicolumn{1}{c}{\emph{RTE} $\downarrow$}\\
			& $\tau_2$ = 5\% & $\tau_1$ = 1cm  & 10mm & 5mm & 2mm & ($^\circ$) & (mm) \\
			\hline
			0 & \textbf{97.3}  & 94.8 & 95.3 & 93.7 & 80.1 & 0.197 & 1.28 \\
			0.5 & 96.8  & 94.6 & 98.4 & 95.7 & 82.3 & 0.192 & 1.24 \\
			\rowcolor{gray!20} 1 & 97.1 & \textbf{95.0}  & \textbf{99.0} & \textbf{96.9} & 83.6 & \textbf{0.190} & \textbf{1.21} \\
			2 & 96.9  & 94.4 & 98.2 & 96.7 & \textbf{83.8} & \textbf{0.190} & 1.22 \\
			\hline
		\end{tabular}
	}
\end{table}

\begin{figure}[t]
	\centering
	\includegraphics[width=0.49\textwidth]{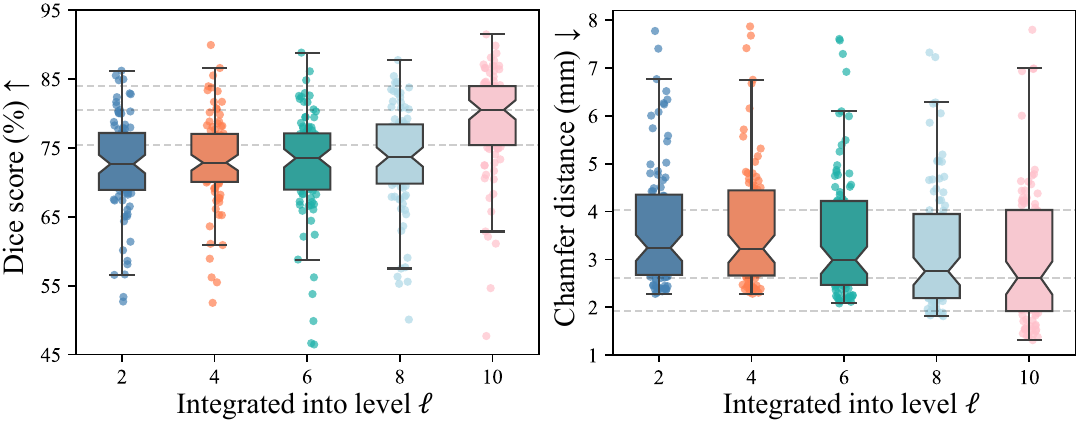}
	\caption{
		Performance of embedding the deformation matrix $\mathfrak{S}$ at different pyramid levels on proposed \ourdata~test set.
	}
	\label{fig:ab_stucture_adaptation}
\end{figure}

\vspace{-10pt}
\subsection{Ablation Studies}\label{abl-study}


\subsubsection{Rigid registration}

\textbf{Different training strategies study:} 
During training, we first train the rigid registration network $\Phi$ on our synthetic data (Sec.~\ref{sec:Dataset}), then freeze parameters of $\Phi$ and continue training our entire framework on \ourdata~until convergence. 
To validate the training scheme, we ablate another strategy where the parameters of $\Phi$ remain unfrozen for further tuning on~\ourdata~after pre-training.
As shown in~\tabref{table:ab_rigid_training_strategy}, not freezing $\Phi$ leads to performance degradation on IR, RR, and Dice metrics, along with increased training time. 
Conversely, the ``frozen" strategy not only achieves shorter training time but also yields superior results. 
It suggests that the domain gap between in-vivo and synthetic data, as well as the limited data size, makes freezing parameters of $\Phi$ beneficial by reducing model complexity and the risk of overfitting, thereby stabilizing the training process.

\textbf{Effect of loss weight:} 
We investigate the impact of loss weights (i.e., $\lambda_{tran}$) on model performance when training the network $\Phi$ using $\mathcal{L}_{corr}$ and $\mathcal{L}_{tran}$.
The results in~\tabref{table:ab_loss_weight_rigid} show that the registration recall can be further improved by incorporating with $\mathcal{L}_{tran}$. 
Setting $\lambda_{tran}$ to 1 yields the best performance in terms of the IR, RR, RRE, and RTE metrics.

\subsubsection{Non-rigid registration} 
\textbf{Effect of structure-guided shape adaptation:} 
Deforming the preoperative model based on the visible liver anatomy is crucial for preoperative-to-intraoperative registration. 
Aside from the qualitative results in ~\figref{fig: main_vis},
we also verify the impact of embedding the adaptation matrix $\mathfrak{S}$ at various levels $\ell$ to validate our shape adaptation strategy.
As shown in~\figref{fig:ab_stucture_adaptation}, there is a notable improvement in Dice and CD metrics at higher frequencies. 
Embedding $\mathfrak{S}$ at the highest frequency yields optimal model performance, as our shape adaptation strategy enhances constraint and guidance over the entire deformation field by measuring the similarity between preoperative and intraoperative liver models.

\textbf{Effect of scale-consistent liver recovery:}
Our model ensure the consistency of scale between the preoperative and intraoperative liver models via the PCA-based scheme. 
To validate the effects of this strategy, we ablate the proposed scale adjustment strategy and compare it with the prevailing pose-scale joint estimation algorithm, \ie, Umeyama~\cite{umeyama1991least}. 
As shown in~\tabref{table:ab_scale_consistency_strategy}, the performance of~\ourmodel~decreases significantly if the proposed scale consistency strategy is not used.
Even using the Umeyama algorithm for concurrent scale and pose estimation, it still performs remarkably lower than~\ourmodel~with scale consistency, \ie, 78.89$\pm$6.76 \emph{vs.} 53.57$\pm$14.22 in Dice and 2.97$\pm$1.06 \emph{vs.} 3.58$\pm$2.09 in CD.

\begin{figure}[t]
	\centering
	\includegraphics[width=0.48\textwidth]{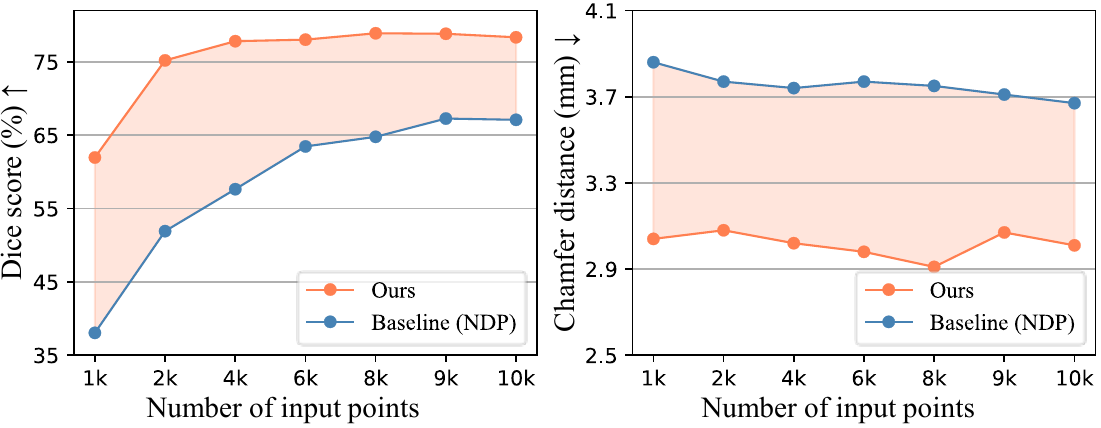}
	\caption{
		Effect of the number of input points on~\ourmodel. 
	}
	\label{fig:ab_different_input_pts}
\end{figure}

\begin{table}[t]
	\footnotesize
	\centering
	\caption{
		Effect of scale-consistent liver point cloud recovery.
	}
	\label{table:ab_scale_consistency_strategy}
		\renewcommand{\arraystretch}{1.1}
		\renewcommand{\tabcolsep}{1.2mm}
		\begin{tabular}{l|c|c|c}
			\hline
			Metrics& \multicolumn{1}{c|}{\makecell{Umeyama~\cite{umeyama1991least}}} & \multicolumn{1}{c|}{\makecell{w/o \\ scale-consistency}} & \multicolumn{1}{c}{\makecell{w/ \\ scale-consistency}} \\
			\hline
			Dice (\%) $\uparrow$ & 53.57$\pm$14.22 & 49.13$\pm$11.83 & \textbf{78.89$\pm$6.76}  \\
			CD (mm) $\downarrow$ & 3.58$\pm$2.09 & 10.66$\pm$4.08 & \textbf{2.97$\pm$1.06}   \\
			\hline
		\end{tabular}
\end{table}

\textbf{Effects of number of input points:}
Different numbers of points of the input liver model may affect feature representation learning, and consequently, the accuracy of predicted pose parameters.  
We further analyze the performance of our model against the baseline model (NDP~\cite{li2022non}) by varying the number of input points. 
The quantitative results in~\figref{fig:ab_different_input_pts} show that as the number of input points increases, the Dice score progressively improves and then stabilizes. 
When the number of input points exceeds 2k, our framework can achieve a satisfactory Dice value. 
The CD metric remains relatively stable, exhibiting only a slight decline, indicating the robustness of our model.
Compared to the baseline, our method demonstrates significant improvement in both Dice and CD metrics.


\textbf{Impact of liver segmentation masks:} Our model operates based on the segmented liver mask, as illustrated in Fig. \ref{fig: pipeline}. To evaluate the impact of varying segmentation results on registration performance, we ablate different segmentation models to automatically produce the liver mask, \ie, SAM~\cite{kirillov2023segment}, SAM-HQ~\cite{ke2023segment}, and MedSAM~\cite{ma2024segment}. In Tab. \ref{tab:varying_seg_depth}, we present a quantitative comparison between the manual annotations and three representative segmentation models. Compared to manual annotations, the predicted masks from deep models achieve comparable yet slightly lower performance, with a maximum drop of 3.56\% in Dice score and 0.15 mm in CD value. It indicates that while the quality of the segmentation mask can affect registration performance to some extent, the overall performance remains stable, demonstrating the robustness of our model to variations in segmentation results.

\textbf{Performance with varying depth estimators: Performance with varying depth estimators: The quality of the estimated depth map may influence the performance of intraoperative liver reconstruction, thereby influencing the registration accuracy. To this end, we conduct an ablation study with three cutting-edge monocular depth estimators in our framework, i.e., DPT~\cite{ranftl2021vision}, MiDaS~\cite{Ranftl2022}, and Depth Anything V2~\cite{yang2024depthv2}, to validate the impact on registration accuracy.
	As summarized in Tab. \ref{tab:varying_seg_depth}, the advanced depth estimation model can further benefit the registration performance of our framework. Concretely, Depth Anything V2 achieves slight improvements in both metrics, while other methods (DPT and MiDaS) exhibit slight performance degradations compared to our default setting, i.e., 75.71\% and 76.13\% vs. 78.89\% in Dice score, and 3.22mm and 3.09mm vs. 2.97mm in CD metric. These results substantiate the robustness of our \ourmodel~framework to variations in depth estimators and highlight its generalizability in accommodating different depth estimation models.
}

\begin{table}[!t]
	\setlength{\tabcolsep}{1.1pt}
	\centering
	\caption{Quantitative results of varying mask segmentation models and depth estimators on \ourdata~dataset.
	}
	\label{tab:varying_seg_depth}
	\setlength{\tabcolsep}{10pt}
	\resizebox{0.97\linewidth}{!}
	{ 
		\begin{tabular}{l|cc}
			\hline
			Models  & DICE (\%) $\uparrow$   & CD (mm) $\downarrow$     
			\cr\hline
			\hline
			\multicolumn{3}{c}{Varying mask segmentation models} 
			\cr\hline
			\hline
			
			\ourcell Manual labeling~(default) & \ourcell \textbf{78.89}\ci{6.76} & \ourcell \textbf{2.97}\ci{1.06}\\
			SAM (ICCV'23)~\cite{kirillov2023segment} & 75.33\ci{6.92} & 3.12\ci{1.69} \\
			SAM-HQ (NeurIPS'23)~\cite{ke2023segment} & 76.91\ci{7.51} & 2.95\ci{1.37} \\
			MedSAM (Nat. Commun.'24)~\cite{ma2024segment} & 76.16\ci{7.39} & 3.01\ci{1.54} 
			\cr\hline
			\hline
			\multicolumn{3}{c}{Varying depth estimation models} 
			\cr\hline
			\hline
			\ourcell Depth Anything V1~\cite{yang2024depth} (default) & \ourcell 78.89\ci{6.76} & \ourcell 2.97\ci{1.06}\\
			DPT (ICCV'21)~\cite{ranftl2021vision} & 75.71\ci{8.48} & 3.22\ci{1.46} \\
			MiDaS (TPAMI'22)~\cite{Ranftl2022} & 76.13\ci{7.45} & 3.09\ci{1.16} \\
			Depth Anything V2 (NeurIPS'24)~\cite{yang2024depthv2} & \textbf{78.94}\ci{6.16} & \textbf{2.91}\ci{1.09} \\
			

			\hline

		\end{tabular}
	}
\end{table}

\subsubsection{Robustness to noise}

\begin{figure}[t]
	\centering
	\includegraphics[width=0.49\textwidth]{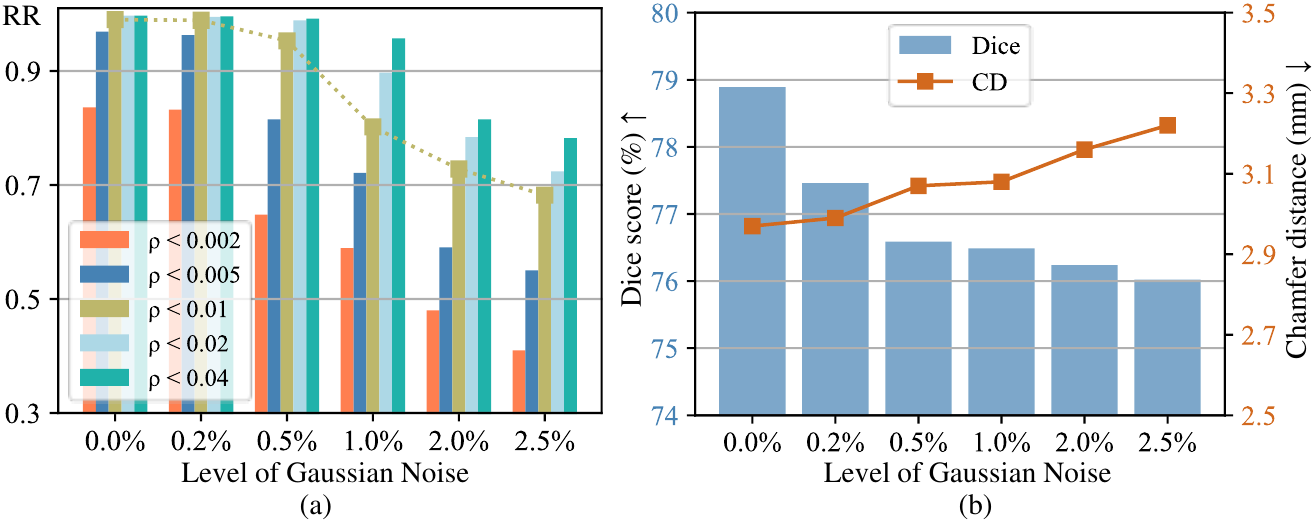}
	\caption{
		Performance of~\ourmodel~under varying Gaussian noise on both synthetic and in-vivo datasets.
	}
	\label{fig:ab_different_gaussian_noise}
\end{figure}

To evaluate the robustness of the proposed \ourmodel~against reconstruction noise during intraoperative liver modeling, we add additional Gaussian noise to the intraoperative liver model with standard deviations ranging from 0.2\% to 2.5\% of the range of the reconstructed point cloud. Fig. \ref{fig:ab_different_gaussian_noise} (a) presents the RR metric under different noise levels and thresholds $\rho$, evaluated on the synthetic dataset. As illustrated, RR gradually declines as noise increases. Notably, under the clinically acceptable threshold ($\rho < 0.01m$) for resection procedures, our model maintains high accuracy (95.3\%) for noise levels up to 0.5\%.
	We further provide quantitative results evaluated on the proposed \ourdata~in-vivo dataset in Fig. \ref{fig:ab_different_gaussian_noise} (b). Specifically, as the noise level increases from 0.2\% to 2.5\%, the Dice score decreases by only 2.87\%, while the CD value shows a marginal increase of 0.25mm. This minor performance degradation highlights the resilience of our method to noise introduced during intraoperative point cloud reconstruction. Even at the highest noise level (2.5\%), the Dice score remains within an acceptable range, and the CD metric does not exhibit a significant rise. The experimental results demonstrate the strong robustness of our framework against intraoperative point cloud perturbations, ensuring reliable registration performance even in the presence of reconstruction noise.

\subsubsection{Impact of different surface visibility}
\begin{figure}[t]
	\centering
	\includegraphics[width=0.49\textwidth]{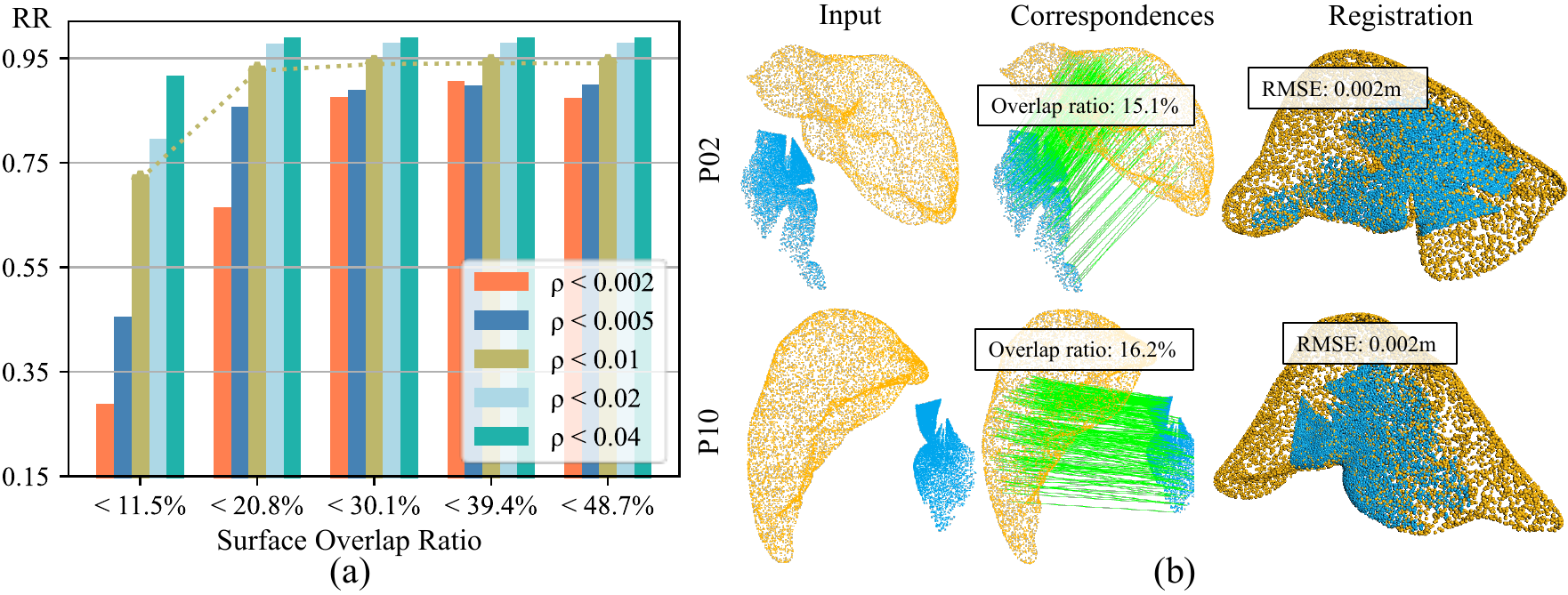}
	\caption{
		Quantitative and qualitative results of~\ourmodel~under varying surface overlap ratios on synthetic data.
	}
	\label{fig:ab_different_overlap_ratio_syn}
\end{figure}

\begin{figure}[t]
	\centering
	\includegraphics[width=0.49\textwidth]{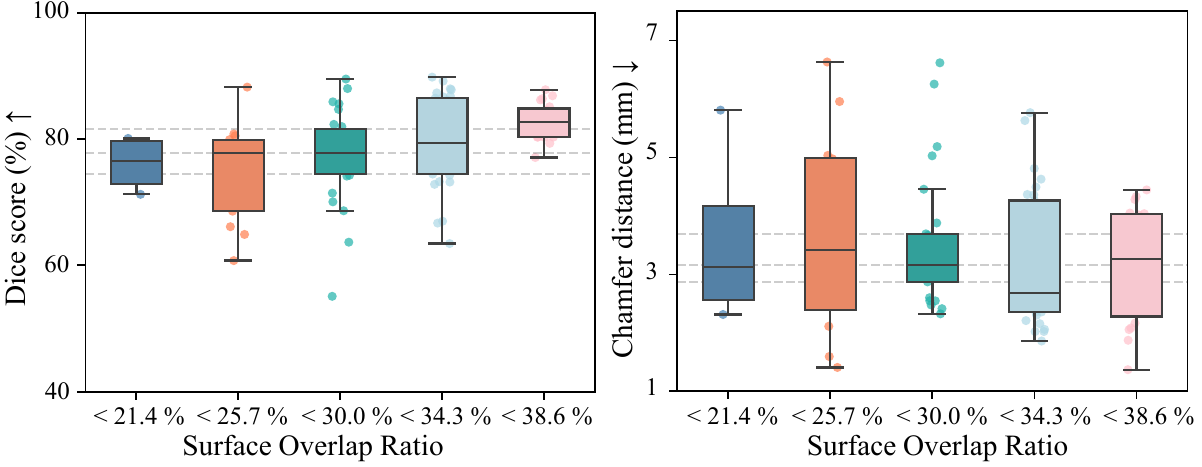}
	\caption{
		Performance of~\ourmodel~under varying surface overlap ratios on proposed~\ourdata~dataset.
	}
	\label{fig:ab_different_overlap_ratio_invivo}
\end{figure}

To investigate the effect of intraoperative liver visibility conditions on registration performance, we further quantify the performance variation of our model under varying surface overlap ratios, defined as the percentage of the liver surface covered by the intraoperative point cloud.
	This overlap ratio reflects various occlusion and visibility conditions during surgery, where a lower ratio indicates higher occlusion or lower visibility of the liver surface.
	Fig.~\ref{fig:ab_different_overlap_ratio_syn} (a) gives the quantitative results of the RR metric with different thresholds $\rho$ on the synthetic data. We can observe that when the overlap ratio falls within the range of 20.8\% to 48.7\%, the registration performance remains high and stable, with an average RR of 94.1\% under the threshold $\rho<0.01m$.
	Notably, even under extremely low overlap ratios (11.5\%-20.8\%), our method still achieves a high registration accuracy (93\%), demonstrating its superiority and robustness to occlusion. We also present the qualitative results in Fig.~\ref{fig:ab_different_overlap_ratio_syn} (b), illustrating that our model consistently achieves reliable registration even under low overlap ratios, demonstrating its effectiveness in challenging visibility conditions.
	We further provide the quantified results based on the varying overall ratios on our proposed \ourdata~dataset in Fig.~\ref{fig:ab_different_overlap_ratio_invivo}. We can see that our method shows consistent stability in both Dice score and CD metrics across different overlap ratios. Even when the overlap ratio falls below 21.4\%, our model achieves nearly 75\% and 3mm in Dice and CD metrics respectively, highlighting its robustness under limited visibility conditions in real-world intraoperative scenarios.

\section{Discussion}
The difficulty in defining and detecting liver landmarks with insufficient in-vivo 3D-2D landmark matching training data, limits the learning capacity and registration performance of landmark-based models. Thus, it is essential to establish a more comprehensive in-vivo preoperative-to-intraoperative dataset and to develop landmark-free registration architectures for improving registration accuracy.

To this end, we propose a task-specific, self-supervised learning framework tailored for liver registration.
	By converting the conventional 3D-2D registration workflow into a 3D-3D point cloud registration pipeline, our method effectively addresses the challenges of partial visibility and deformation in intraoperative scenarios. 
	
	The proposed feature disentangled transformer empowers our model to learn effective point correspondences, enabling accurate rigid transformation estimation even without landmarks.
With the ``frozen" training strategy, \ourmodel~achieves improved performance in both rigid and non-rigid registration and reduced training times.
For non-rigid deformation, we introduce a structure-regularized deformation network that models structural correlations through geometry similarity learning in a low-rank transformer network, offering coarse-to-fine deformable alignment. 
	This strategy allows us to fully exploit the spatial features of the intraoperative liver model, mitigating inconsistencies caused by relying solely on local features and ensuring spatially coherent deformation across the liver surface.
Additionally, we contribute a new in-vivo registration dataset, \ourdata, which contains more patient cases, keyframes, auxiliary 2D liver masks, and 3D point clouds than existing datasets, facilitating performance verification and development of registration frameworks.

Despite our pioneering attempt at a 3D-3D liver registration pipeline, there are some limitations and schemes that need to be optimized. 

First, scale-consistent calibration is currently performed as an offline preprocessing step using geometric assumptions. Although this approach provides stable initialization, it is not integrated fully into the network optimization.
	While preliminary attempts have been made to incorporate scale estimation into the registration pipeline, these efforts have not led to improved performance.
	We will further explore joint-scale optimization strategies to leverage self-supervised learning more effectively in the future work.
Second, due to the inherent uncertainty of intraoperative liver deformation, our framework ensures precise fitting only for visible portions, while using a relatively static approach for invisible areas. 
To bridge this gap, we intend to generate additional training data by simulating respiratory effects and pneumoperitoneum with physics-based models.

\section{Conclusion}
In this work, we introduce \ourmodel, a novel landmark-free preoperative-to-intraoperative registration model with effective self-supervised learning.
\ourmodel~reformulates 3D-2D registration as a 3D-3D point cloud registration task, decoupling it into rigid and non-rigid registration steps. 
In~\ourmodel, a feature-disentangled transformer followed by a structure-regularized deformation network is designed to learn robust correspondences and deformation parameters. 
Extensive experiments on both synthetic and our in-vivo datasets demonstrate that our method achieves superior performance than existing methods in both rigid and non-rigid registration perspectives. User studies on~\ourdata~dataset also prove that our method meets the clinical needs of surgeons.





\bibliographystyle{IEEEtran}
\bibliography{ref}

\begin{thebibliography}{10}
\providecommand{\url}[1]{#1}
\csname url@samestyle\endcsname
\providecommand{\newblock}{\relax}
\providecommand{\bibinfo}[2]{#2}
\providecommand{\BIBentrySTDinterwordspacing}{\spaceskip=0pt\relax}
\providecommand{\BIBentryALTinterwordstretchfactor}{4}
\providecommand{\BIBentryALTinterwordspacing}{\spaceskip=\fontdimen2\font plus
\BIBentryALTinterwordstretchfactor\fontdimen3\font minus
  \fontdimen4\font\relax}
\providecommand{\BIBforeignlanguage}[2]{{%
\expandafter\ifx\csname l@#1\endcsname\relax
\typeout{** WARNING: IEEEtran.bst: No hyphenation pattern has been}%
\typeout{** loaded for the language `#1'. Using the pattern for}%
\typeout{** the default language instead.}%
\else
\language=\csname l@#1\endcsname
\fi
#2}}
\providecommand{\BIBdecl}{\relax}
\BIBdecl

\bibitem{ramalhinho2023value}
J.~Ramalhinho, S.~Yoo, T.~Dowrick, B.~Koo, M.~Somasundaram, K.~Gurusamy, D.~J.
  Hawkes, B.~Davidson, A.~Blandford, and M.~J. Clarkson, ``The value of
  augmented reality in surgery—a usability study on laparoscopic liver
  surgery,'' \emph{MedIA}, vol.~90, p. 102943, 2023.

\bibitem{guo2025surgical}
D.~Guo, W.~Si, Z.~Li, J.~Pei, and P.-A. Heng, ``Surgical workflow recognition
  and blocking effectiveness detection in laparoscopic liver resection with
  pringle maneuver,'' in \emph{AAAI}, vol.~39, no.~3, 2025, pp. 3220--3228.

\bibitem{koo2022automatic}
B.~Koo, M.~R. Robu, M.~Allam, M.~Pfeiffer, S.~Thompson, K.~Gurusamy,
  B.~Davidson, S.~Speidel, D.~Hawkes \emph{et~al.}, ``Automatic, global
  registration in laparoscopic liver surgery,'' \emph{IJCARS}, pp. 1--10, 2022.

\bibitem{mhiri2024neural}
I.~Mhiri, D.~Pizarro, and A.~Bartoli, ``Neural patient-specific 3d--2d
  registration in laparoscopic liver resection,'' \emph{IJCARS}, pp. 1--8,
  2024.

\bibitem{pelanis2021evaluation}
E.~Pelanis, A.~Teatini, B.~Eigl, A.~Regensburger, A.~Alzaga, R.~P. Kumar,
  T.~Rudolph, D.~L. Aghayan, C.~Riediger, N.~Kvarnstr{\"o}m \emph{et~al.},
  ``Evaluation of a novel navigation platform for laparoscopic liver surgery
  with organ deformation compensation using injected fiducials,'' \emph{MedIA},
  vol.~69, p. 101946, 2021.

\bibitem{ozgur2018preoperative}
E.~{\"O}zg{\"u}r, B.~Koo, B.~Le~Roy \emph{et~al.}, ``Preoperative liver
  registration for augmented monocular laparoscopy using backward--forward
  biomechanical simulation,'' \emph{IJCARS}, vol.~13, pp. 1629--1640, 2018.

\bibitem{feuerstein2008intraoperative}
M.~Feuerstein, T.~Mussack, S.~M. Heining, and N.~Navab, ``Intraoperative
  laparoscope augmentation for port placement and resection planning in
  minimally invasive liver resection,'' \emph{IEEE Transactions on Medical
  Imaging}, vol.~27, no.~3, pp. 355--369, 2008.

\bibitem{koo2017deformable}
B.~Koo, E.~{\"O}zg{\"u}r, B.~Le~Roy, E.~Buc, and A.~Bartoli, ``Deformable
  registration of a preoperative 3d liver volume to a laparoscopy image using
  contour and shading cues,'' in \emph{MICCAI}, 2017, pp. 326--334.

\bibitem{espinel2022using}
Y.~Espinel, L.~Calvet, K.~Botros, E.~Buc, C.~Tilmant, and A.~Bartoli, ``Using
  multiple images and contours for deformable 3d--2d registration of a
  preoperative ct in laparoscopic liver surgery,'' \emph{IJCARS}, vol.~17,
  no.~12, pp. 2211--2219, 2022.

\bibitem{pei2024land}
J.~Pei, R.~Cui, Y.~Li, W.~Si, J.~Qin, and P.-A. Heng, ``Depth-driven geometric
  prompt learning for laparoscopic liver landmark detection,'' in
  \emph{MICCAI}, 2024.

\bibitem{labrunie2023automatic}
M.~Labrunie, D.~Pizarro, C.~Tilmant, and A.~Bartoli, ``Automatic 3d/2d
  deformable registration in minimally invasive liver resection using a mesh
  recovery network.'' in \emph{MIDL}, 2023, pp. 1104--1123.

\bibitem{zhang2019toward}
Y.~Zhang, F.~Li, L.~Qiu, L.~Xu, X.~Niu, Y.~Sui, S.~Zhang, Q.~Zhang, and
  L.~Zhang, ``Toward precise osteotomies: a coarse-to-fine 3d cut plane
  planning method for image-guided pelvis tumor resection surgery,'' \emph{IEEE
  Transactions on Medical Imaging}, vol.~39, no.~5, pp. 1511--1523, 2019.

\bibitem{espinel2024keyhole}
Y.~Espinel, N.~Rabbani, T.~B. Bui, M.~Ribeiro, E.~Buc, and A.~Bartoli,
  ``Keyhole-aware laparoscopic augmented reality,'' \emph{MedIA}, vol.~94, p.
  103161, 2024.

\bibitem{lei2024epicardium}
L.~Lei, J.~Zhou, J.~Pei, B.~Zhao, Y.~Jin, Y.-C.~J. Teoh, J.~Qin, and P.-A.
  Heng, ``Epicardium prompt-guided real-time cardiac ultrasound frame-to-volume
  registration,'' in \emph{MICCAI}, 2024.

\bibitem{pfeiffer2020non}
M.~Pfeiffer, C.~Riediger, S.~Leger, J.-P. K{\"u}hn, D.~Seppelt, R.-T. Hoffmann,
  J.~Weitz \emph{et~al.}, ``Non-rigid volume to surface registration using a
  data-driven biomechanical model,'' in \emph{MICCAI}, 2020, pp. 724--734.

\bibitem{collins2020augmented}
T.~Collins, D.~Pizarro, S.~Gasparini, N.~Bourdel, P.~Chauvet, M.~Canis,
  L.~Calvet, and A.~Bartoli, ``Augmented reality guided laparoscopic surgery of
  the uterus,'' \emph{IEEE Transactions on Medical Imaging}, vol.~40, no.~1,
  pp. 371--380, 2020.

\bibitem{wang2024video}
E.~Wang, Y.~Liu, P.~Tu, Z.~A. Taylor \emph{et~al.}, ``Video-based soft tissue
  deformation tracking for laparoscopic augmented reality-based navigation in
  kidney surgery,'' \emph{IEEE Transactions on Medical Imaging}, 2024.

\bibitem{zhang2024point}
Y.~Zhang, Y.~Zou, and P.~X. Liu, ``Point cloud registration in laparoscopic
  liver surgery using keypoint correspondence registration network,''
  \emph{IEEE Transactions on Medical Imaging}, 2024.

\bibitem{guo2013rotational}
Y.~Guo, F.~Sohel, M.~Bennamoun, M.~Lu, and J.~Wan, ``Rotational projection
  statistics for 3d local surface description and object recognition,''
  \emph{IJCV}, vol. 105, pp. 63--86, 2013.

\bibitem{rusinkiewicz2019symmetric}
S.~Rusinkiewicz, ``A symmetric objective function for icp,'' \emph{ACM TOG},
  vol.~38, no.~4, pp. 1--7, 2019.

\bibitem{yew2022regtr}
Z.~J. Yew and G.~H. Lee, ``Regtr: End-to-end point cloud correspondences with
  transformers,'' in \emph{IEEE CVPR}, 2022, pp. 6677--6686.

\bibitem{huang2021predator}
S.~Huang, Z.~Gojcic, M.~Usvyatsov, A.~Wieser, and K.~Schindler, ``Predator:
  Registration of 3d point clouds with low overlap,'' in \emph{IEEE CVPR},
  2021, pp. 4267--4276.

\bibitem{li2022lepard}
Y.~Li and T.~Harada, ``Lepard: Learning partial point cloud matching in rigid
  and deformable scenes,'' in \emph{IEEE CVPR}, 2022, pp. 5554--5564.

\bibitem{qin2023geotransformer}
Z.~Qin, H.~Yu, C.~Wang, Y.~Guo, Y.~Peng, S.~Ilic, D.~Hu, and K.~Xu,
  ``Geotransformer: Fast and robust point cloud registration with geometric
  transformer,'' \emph{IEEE TPAMI}, vol.~45, no.~8, pp. 9806--9821, 2023.

\bibitem{li2008global}
H.~Li, R.~W. Sumner, and M.~Pauly, ``Global correspondence optimization for
  non-rigid registration of depth scans,'' in \emph{Computer Graphics Forum},
  vol.~27, no.~5, 2008, pp. 1421--1430.

\bibitem{bozic2021neural}
A.~Bozic, P.~Palafox, M.~Zollhofer, J.~Thies, A.~Dai, and M.~Nie{\ss}ner,
  ``Neural deformation graphs for globally-consistent non-rigid
  reconstruction,'' in \emph{IEEE CVPR}, 2021, pp. 1450--1459.

\bibitem{li2022non}
Y.~Li and T.~Harada, ``Non-rigid point cloud registration with neural
  deformation pyramid,'' \emph{NeurIPS}, vol.~35, pp. 27\,757--27\,768, 2022.

\bibitem{vaswani2017attention}
A.~Vaswani, N.~Shazeer, N.~Parmar, J.~Uszkoreit, L.~Jones, A.~N. Gomez,
  {\L}.~Kaiser \emph{et~al.}, ``Attention is all you need,'' \emph{NeurIPS},
  vol.~30, 2017.

\bibitem{ALI2024103371}
S.~Ali, Y.~Espinel, Y.~Jin, P.~Liu, B.~G{\"u}ttner, X.~Zhang, L.~Zhang,
  T.~Dowrick, M.~J. Clarkson, S.~Xiao \emph{et~al.}, ``An objective comparison
  of methods for augmented reality in laparoscopic liver resection by
  preoperative-to-intraoperative image fusion from the miccai2022 challenge,''
  \emph{MedIA}, p. 103371, 2024.

\bibitem{zhang2000flexible}
Z.~Zhang, ``A flexible new technique for camera calibration,'' \emph{IEEE
  TPAMI}, vol.~22, no.~11, pp. 1330--1334, 2000.

\bibitem{pfeiffer2019generating}
M.~Pfeiffer, I.~Funke, M.~R. Robu, S.~Bodenstedt, L.~Strenger, S.~Engelhardt,
  T.~Ro{\ss}, M.~J. Clarkson, K.~Gurusamy \emph{et~al.}, ``Generating large
  labeled data sets for laparoscopic image processing tasks using unpaired
  image-to-image translation,'' in \emph{MICCAI}, 2019, pp. 119--127.

\bibitem{besl1992method}
P.~J. Besl and N.~D. McKay, ``Method for registration of 3-d shapes,'' in
  \emph{Sensor Fusion IV: control paradigms and data structures}, vol. 1611,
  1992, pp. 586--606.

\bibitem{yang2024depth}
L.~Yang, B.~Kang, Z.~Huang, X.~Xu, J.~Feng, and H.~Zhao, ``Depth anything:
  Unleashing the power of large-scale unlabeled data,'' in \emph{IEEE CVPR},
  2024, pp. 10\,371--10\,381.

\bibitem{wold1987principal}
S.~Wold, K.~Esbensen, and P.~Geladi, ``Principal component analysis,''
  \emph{Chemometrics and Intelligent Laboratory Systems}, vol.~2, no. 1-3, pp.
  37--52, 1987.

\bibitem{thomas2019kpconv}
H.~Thomas, C.~R. Qi, J.-E. Deschaud, B.~Marcotegui, F.~Goulette, and L.~J.
  Guibas, ``Kpconv: Flexible and deformable convolution for point clouds,'' in
  \emph{IEEE ICCV}, 2019, pp. 6411--6420.

\bibitem{pei2024s}
J.~Pei, D.~Guo, J.~Zhang, M.~Lin, Y.~Jin, and P.-A. Heng, ``S2former-or:
  Single-stage bi-modal transformer for scene graph generation in or,''
  \emph{IEEE Transactions on Medical Imaging}, 2024.

\bibitem{qi2017pointnet++}
C.~R. Qi, L.~Yi, H.~Su, and L.~J. Guibas, ``Pointnet++: Deep hierarchical
  feature learning on point sets in a metric space,'' \emph{NeurIPS}, vol.~30,
  2017.

\bibitem{wang2020linformer}
S.~Wang, B.~Z. Li, M.~Khabsa, H.~Fang \emph{et~al.}, ``Linformer:
  Self-attention with linear complexity,'' \emph{arXiv preprint
  arXiv:2006.04768}, 2020.

\bibitem{fischler1981random}
M.~A. Fischler and R.~C. Bolles, ``Random sample consensus: a paradigm for
  model fitting with applications to image analysis and automated
  cartography,'' \emph{Communications of the ACM}, vol.~24, no.~6, pp.
  381--395, 1981.

\bibitem{ravi2020pytorch3d}
N.~Ravi, J.~Reizenstein, D.~Novotny, T.~Gordon, W.-Y. Lo, J.~Johnson, and
  G.~Gkioxari, ``Accelerating 3d deep learning with pytorch3d,''
  \emph{arXiv:2007.08501}, 2020.

\bibitem{feydy2019interpolating}
J.~Feydy, T.~S{\'e}journ{\'e}, F.-X. Vialard, S.-i. Amari, A.~Trouv{\'e}, and
  G.~Peyr{\'e}, ``Interpolating between optimal transport and mmd using
  sinkhorn divergences,'' in \emph{AISTATS}, 2019, pp. 2681--2690.

\bibitem{park2021nerfies}
K.~Park, U.~Sinha, J.~T. Barron, S.~Bouaziz, D.~B. Goldman, S.~M. Seitz, and
  R.~Martin-Brualla, ``Nerfies: Deformable neural radiance fields,'' in
  \emph{IEEE ICCV}, 2021, pp. 5865--5874.

\bibitem{li2021neural}
X.~Li, J.~Kaesemodel~Pontes, and S.~Lucey, ``Neural scene flow prior,''
  \emph{NeurIPS}, vol.~34, pp. 7838--7851, 2021.

\bibitem{Prokudin_2023_ICCV}
S.~Prokudin, Q.~Ma, M.~Raafat, J.~Valentin, and S.~Tang, ``Dynamic point
  fields,'' in \emph{IEEE ICCV}, October 2023, pp. 7964--7976.

\bibitem{zhao2024clustereg}
M.~Zhao, J.~Jiang, L.~Ma, S.~Xin, G.~Meng, and D.-M. Yan, ``Correspondence-free
  non-rigid point set registration using unsupervised clustering analysis,'' in
  \emph{IEEE CVPR}, 2024, pp. 21\,199--21\,208.

\bibitem{umeyama1991least}
S.~Umeyama, ``Least-squares estimation of transformation parameters between two
  point patterns,'' \emph{IEEE TPAMI}, vol.~13, no.~04, pp. 376--380, 1991.

\bibitem{kirillov2023segment}
A.~Kirillov, E.~Mintun, N.~Ravi, H.~Mao, C.~Rolland, L.~Gustafson, T.~Xiao,
  S.~Whitehead, A.~C. Berg, W.-Y. Lo \emph{et~al.}, ``Segment anything,'' in
  \emph{IEEE ICCV}, 2023, pp. 4015--4026.

\bibitem{ke2023segment}
L.~Ke, M.~Ye, M.~Danelljan, Y.-W. Tai, C.-K. Tang, F.~Yu \emph{et~al.},
  ``Segment anything in high quality,'' \emph{NeurIPS}, vol.~36, pp.
  29\,914--29\,934, 2023.

\bibitem{ma2024segment}
J.~Ma, Y.~He, F.~Li, L.~Han, C.~You, and B.~Wang, ``Segment anything in medical
  images,'' \emph{Nature Communications}, vol.~15, no.~1, p. 654, 2024.

\bibitem{ranftl2021vision}
R.~Ranftl, A.~Bochkovskiy, and V.~Koltun, ``Vision transformers for dense
  prediction,'' in \emph{IEEE ICCV}, 2021, pp. 12\,179--12\,188.

\bibitem{Ranftl2022}
R.~Ranftl, K.~Lasinger, D.~Hafner, K.~Schindler, and V.~Koltun, ``Towards
  robust monocular depth estimation: Mixing datasets for zero-shot
  cross-dataset transfer,'' \emph{IEEE TPAMI}, vol.~44, no.~3, 2022.

\bibitem{yang2024depthv2}
L.~Yang, B.~Kang, Z.~Huang, Z.~Zhao, X.~Xu, J.~Feng, and H.~Zhao, ``Depth
  anything v2,'' \emph{NeurIPS}, vol.~37, pp. 21\,875--21\,911, 2024.

\end{thebibliography}


\end{document}